\documentclass{bmvc2k}

\title{3D-QAE: Fully Quantum Auto-Encoding of 3D Point Clouds}

\addauthor{Lakshika Rathi}{lrathi@wisc.edu}{1,2,$\dagger$}
\addauthor{Edith Tretschk}{edithtretschk.github.io}{2}
\addauthor{Christian Theobalt}{theobalt@mpi-inf.mpg.de}{2}
\addauthor{Rishabh Dabral}{rdabral@mpi-inf.mpg.de}{2}
\addauthor{Vladislav Golyanik}{golyanik@mpi-inf.mpg.de}{2}

\addinstitution{
 Indian Institute of Technology Delhi,\\
 New Delhi, India
}
\addinstitution{
 Max Planck Institute for Informatics,\\
 SIC, Saarbr\"ucken, Germany \\
 \vspace{68pt}
 {\scriptsize $\dagger$work done during an internship at MPI} 
}

\runninghead{Rathi \emph{et al.}}{Fully Quantum Auto-Encoding of 3D Point Clouds}

\usepackage{wrapfig}
\usepackage{gensymb}
\usepackage{breqn}
\usepackage{comment}
\usepackage{multirow}
\usepackage{float}
\usepackage{physics}
\usepackage{amssymb}
\usepackage{enumitem}

\usepackage{color}
\definecolor{bmvcblue}{RGB}{0,24,102}

\begin{document}

\maketitle

\begin{abstract}
    
Existing methods for learning 3D  representations are deep neural networks trained and tested on classical hardware. 
Quantum machine learning architectures, despite their 
theoretically predicted 
advantages in terms of speed and the representational capacity, have so far not been considered for this problem nor for tasks involving 3D data in general. 
This paper thus introduces the first quantum auto-encoder for 3D point clouds. 
Our \emph{3D-QAE} approach is \textit{fully quantum}, \textit{i.e.} all its data processing components are designed for quantum hardware. 
It is trained on collections of 3D point clouds to produce their compressed representations. 
Along with finding a suitable architecture, the core challenges in designing such a fully quantum model include 3D data normalisation and parameter optimisation, and we propose solutions for both these tasks. 
Experiments on simulated gate-based quantum hardware demonstrate that our method outperforms simple classical baselines, paving the way for a new research direction in 3D computer vision. 
The source code is available at \url{https://4dqv.mpi-inf.mpg.de/QAE3D/}.

\end{abstract}

\section{Introduction}\label{sec:intro}

Bolstered by the wide accessibility of experimental quantum hardware and software Quantum Computing (QC) simulators, the emerging fields of Quantum Computer Vision (QCV) 
\cite{golyanik2020quantum, QuantumSync2021, Meli_2022_CVPR, Yang_2022_CVPR, 9879497, Yurtsever2022, Bhatia2023CCuantuMM} 
and Quantum Machine Learning (QML) \cite{Biamonte_2017,Havl_ek_2019,Lloyd_2014,Amin_2018,kwak2021quantum,Abbas_2021,Jerbi_2023,https://doi.org/10.48550/arxiv.2206.03066, SeelbachBenkner2023} have witnessed a steadily growing number of research works.
Most of them are motivated by multiple advantages QC promises over classical computing such as better complexity classes \cite{NielsenChuang2000}, fast convergence, and lower numbers of parameters in machine learning architectures \cite{Biamonte_2017}. 
While most QCV works \cite{QuantumSync2021, SeelbachBenkner2021, Zaech_2022_CVPR, Farina2023quantum} exploit the quantum annealing paradigm of QC that solves quadratic binary unconstrained optimisation objectives only
(which can be too restrictive for many vision problems), 
this paper uses gate-based QC supporting a universal set of operations on qubits. 
We found this flexibility helpful for designing a fully quantum architecture.

Despite this flurry of works, QC and QML algorithms  have not yet been applied to 3D point clouds, which are of core interest in 3D computer vision and graphics communities. 
We start the exploration by looking at auto-encoding 3D point clouds with known correspondences. 
Our motivation stems from existing works on shape auto-encoders \cite{Tretschk2020DEMEA, Zhou2020} and quantum auto-encoders in different scenarios \cite{Romero_2017} suggesting that quantum 3D point cloud auto-encoding should be feasible.

While works on classical mesh auto-encoders \cite{ranjan2018generating,bouritsas2019neural,Tretschk2020DEMEA} often deploy sophisticated components like graph convolutions, they also hint at the strength of a very basic design: flattening the mesh coordinates into a vector, encoding it into a bottleneck latent space via a single fully connected layer followed by an activation function, and then decoding it via another single fully connected layer. 
For example, CoMA \cite{ranjan2018generating} focuses on very small latent spaces in order to show an advantage of its architecture over such a simple baseline. 
Similarly, DEMEA \cite{Tretschk2020DEMEA} only outperforms such baselines when using small latent spaces. 
Since practical quantum computing is in its infancy, sophisticated designs are hardly feasible. 
However, it is possible to design a Quantum Neural Network (QNN) closely resembling a basic classical architecture. 
This makes mesh auto-encoding particularly promising and well-suited for us as it means that we can start the exploration into quantum 3D scene representations with simple quantum architectures.

\begin{wrapfigure}{r}{0.45\columnwidth}
\centering
     \includegraphics[width=0.45\columnwidth]{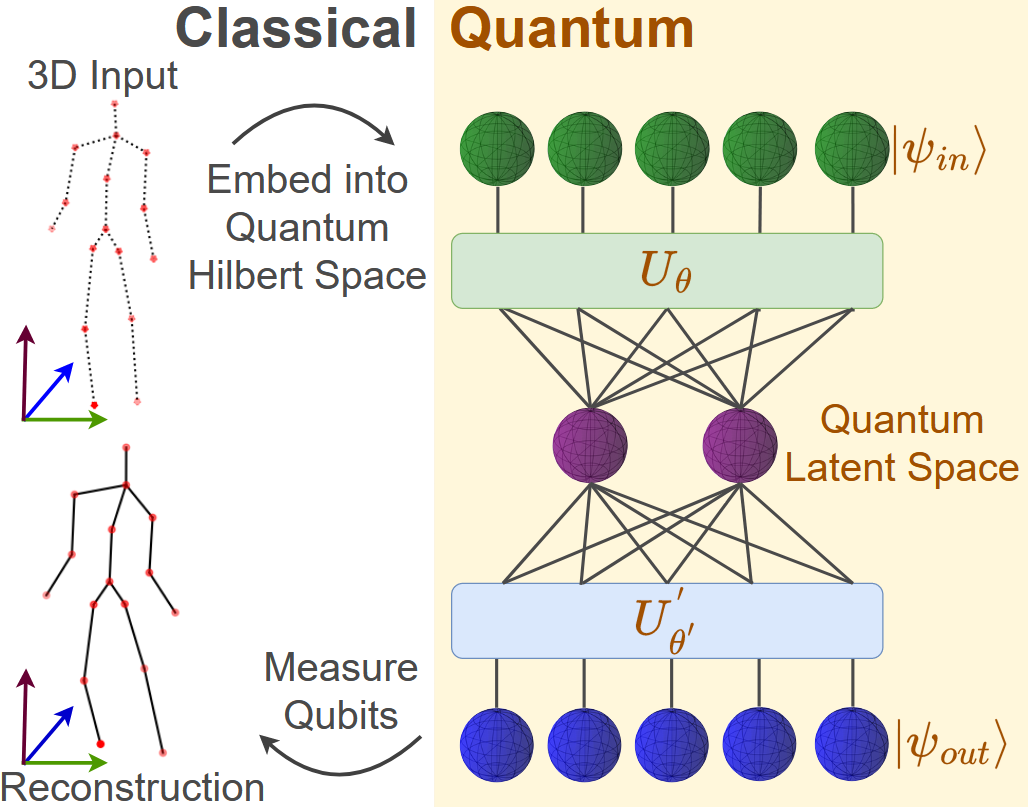}
     \caption{
     \textbf{We use auto-encoding to learn a quantum 3D point cloud representation.}  
     We first embed a classical input into a quantum state $\ket{\psi_{\mathit{in}}}$ of entangled qubits (visualised as spheres), run a learned encoder quantum circuit $U_\theta$ on it, and then apply a quantum non-linearity to obtain the quantum latent  variable. 
     We then run a learned decoder quantum circuit $U'_{\theta'}$ on the latent vector to obtain the output quantum state $\ket{\psi_{\mathit{out}}}$ and 
     finally measure the qubits to recover a classical, reconstructed output 3D point cloud. 
     }
     \label{fig:teaser_fig}
\end{wrapfigure}

Consequently, this paper introduces 3D-QAE, \textit{the first quantum auto-encoder for 3D point sets and investigates its properties}; see Fig.~\ref{fig:teaser_fig}. 
Even though hybrid networks, which combine classical and quantum components, are widespread in the QML literature  \cite{alam2021quantum,sebastianelli2021circuitbased,e25030427,2021arXiv210106189Y,Mari_2020,9885859, Srikumar:2021yzo}, 
we find that classical components dominate the useful processing in hybrid architectures, making up for and hiding the shortcomings of the quantum components. 
This is not surprising, as fully quantum or hybrid architectures are often reported in the literature not to outperform the classical counterparts \cite{Huang2021, Lockwood2021, Lau2022, Bokhan2022}. 
Therefore, to better assess the true potential of quantum computing, we focus on a non-hybrid, \textit{purely quantum method}, \textit{e.g.}~we do not apply classical non-linearities but, instead, employ a fully quantum non-linearity. 
Designing and training fully quantum architectures remains very challenging in general. 
Thus, we need to carefully normalise the data and introduce auxiliary values in order to deal with the input and output restrictions of quantum circuits when using 3D point clouds. 
We conduct experiments on articulated human poses from the AMASS dataset~\cite{AMASS:ICCV:2019} to evaluate our design choices, in particular the circuit architecture, the non-linearity, and the optimisation scheme. 
Summa sumarrum, our primary \textbf{technical contributions} are as follows: 
\begin{itemize}[nolistsep]
    \setlength{\itemsep}{2pt}
    \setlength{\parskip}{0pt}
    \item 3D-QAE, a fully quantum gate-based architecture for 3D point clouds auto-encoding, 
    \item Data normalisation scheme to make point sets compatible with quantum circuits, and 
    \item A quantum gate sequence for improved information propagation after the bottleneck. 
\end{itemize}

\section{Related Work}\label{sec:relatedwork}

\noindent\textbf{Classical Auto-Encoders.}
Auto-encoders were introduced by Rumelhart and McClelland 
\cite{6302929} and have traditionally been used for dimensionality reduction and data analysis \cite{Kramer1991NonlinearPC, article}. 
Classical auto-encoders for meshes and point clouds have been used for a variety of applications \cite{8578506,8578710,tan2018mesh,yang2018foldingnet,Elbaz2017,bouritsas2019neural,hahner2022mesh}. 
A prominent application is learning a latent representation of deformable 3D meshes \cite{yang2018foldingnet,yan2022implicit}. 
Thus, Ranjan \emph{et al.}\ \cite{ranjan2018generating} use spectral graph convolutions \cite{defferrard2016convolutional} with upsampling and downsampling operations to auto-encode meshes of human faces. 

\noindent\textbf{Quantum Machine Learning (QML).} 
The hope that quantum systems can outperform classical systems at identifying atypical patterns in data gave rise to the field of QML \cite{Biamonte_2017}. 
The past years have witnessed the birth of quantum analogues of a range of classical machine learning algorithms, including support vector machines \cite{Havl_ek_2019}, principal component analysis \cite{Lloyd_2014}, and Boltzmann machines \cite{Amin_2018}. 
In particular, \emph{Quantum Neural Networks (QNN)} are the quantum analogue of classical neural networks. 
A QNN consists of parametrised quantum circuits applied on an input quantum state generated by a feature map \cite{kwak2021quantum}. 
Sometimes, QNNs achieve \cite{Abbas_2021} higher expressibility and better trainability than classical counterparts. %

\emph{Quantum Auto-Encoders} (QAE) are a special case of QNNs. 
QAE often follow layered architecture design, 
which allows to use quantum systems in tasks analogous to classical machine learning. 
They are often designed as hybrid systems combining quantum and classical layers to cater to a range of tasks: labeling classical data 
\cite{Mangini_2022}, reconstructing and sampling of biological drugs \cite{li2022scalable}, searching anomalous data points in a given classical dataset \cite{Sakhnenko:2021jme}, and compressing information to enable efficient communication between a client and a server in a quantum cloud computing setting \cite{Zhu:2021jnc}. 
Similarly to how classical auto-encoders learn low-dimensional representations of classical data, quantum auto-encoders \cite{Romero_2017,Srikumar:2021yzo} can be used to represent quantum data in a compressed form: 
Bravo-Prieto \cite{Bravo_Prieto_2021} demonstrates the use of parametrised quantum circuits as variational models to compress Ising models and handwritten digits with high fidelity. 
Effective error detection is crucial for contemporary quantum devices and 
quantum auto-encoders can be efficiently used for detecting and mitigating quantum errors \cite{Zhang_2021} \cite{locher2023quantum}. 
Finally, they have also been shown to effectively denoise 
Greenberger-Horne-Zeilinger 
states from noisy quantum channels \cite{2020arXiv201214714A}.

\noindent\textbf{Quantum Computer Vision and Graphics (QCV/CG).} 
A range of quantum methods have been developed for tasks in computer vision like object detection \cite{8895167} and image classification \cite{alam2021quantum,2021QuIP...20..381C}. 
The limited number of physical qubits in gate-based models has led to a preference for adiabatic quantum computing for most tasks in computer vision \cite{9932103}, such as robust fitting \cite{9879497, Farina2023quantum}, multiple object tracking \cite{electronics10192406}, permutation synchronisation \cite{QuantumSync2021, Yurtsever2022}, and transformation estimation \cite{golyanik2020quantum, Meli_2022_CVPR, SeelbachBenkner2023}. 
However, 
hybrid quantum-classical gate-based models 
are also emerging 
\cite{Yang_2022_CVPR}. 
Our method is designed for the gate-based quantum computing paradigm 
and it is the first to learn 3D point cloud representations with QML.

\begin{figure*}
    \includegraphics[width=\textwidth]{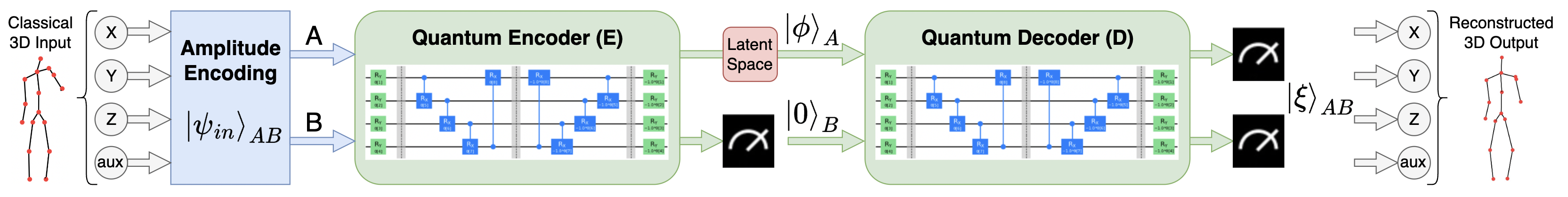}
    \caption{\textbf{3D-QAE, a quantum point cloud auto-encoder.} 
    We prepare a classical 3D point cloud as input and then encode it into a quantum state vector $\ket{\psi_{\mathit{in}}}$ of two sets of qubits, $A$ and $B$, via amplitude encoding. 
    The encoder $E$ (visualised here with $J{=}1$ block) acts on this state vector via a learned unitary transform implemented by a parametrised quantum circuit. 
    At the bottleneck, we remove the information stored in the qubits $B$. 
    This removal acts as a quantum non-linearity whose output is the latent vector $\ket{\phi}$ of qubits $A$. 
    We re-initialise qubits $B$ to 
    $\ket{0}$ 
    and let the decoder $D$, whose architecture is the same as $E$'s, transform qubits $A$ and $B$. 
    We then measure the output of $D$ to obtain the state vector $\ket{\xi}$, which we can classically process in a loss function or convert to the final 3D output reconstruction. 
    }
    \label{fig:full_QA}
\end{figure*}

\section{Method}\label{sec:method}

We propose \textit{3D-QAE}, \textit{i.e.}~a fully quantum, circuit-based auto-encoder for registered 3D point clouds. 
When using current classical hardware to simulate quantum hardware, it scales to dozens of points per point cloud.
Fig.~\ref{fig:full_QA} shows the scheme of 3D-QAE. 
For background on gate-based quantum computing, we refer to the supplementary material.

Like classical auto-encoders, quantum auto-encoders consist of an encoder and a decoder, with a bottleneck latent space in the middle. 
First, the classical data is encoded into a quantum state (Sec.~\ref{sec:input}) that 
is passed to the encoder  (Sec.~\ref{sec:designing_circuit}). 
The encoder output is compressed into a low-dimensional latent space, before being passed to the decoder (Sec.~\ref{sec:architecture}). 
Finally, we optimise for the best parameters of the auto-encoder by encouraging similarity between the decoder output and the input (Sec.~\ref{sec:training}).

\subsection{Input Data Normalisation and Encoding}\label{sec:input}

We take as input a classical 3D point cloud $\{\mathbf{v}_i\in\mathbb{R}^3\}^{V-1}_{i=0}$ with $V$ vertices. 
Several steps are necessary to turn this point cloud into a \textit{state vector} that can be input into a quantum circuit. 
Furthermore, as we discuss later in Sec.~\ref{sec:training}, the output of the auto-encoder necessarily lies in the positive octant since it is a vector $(a_0,\ldots,a_{2^N-1})$ with $a_i\geq 0$ and $\sum_i \abs{a_i}^2 = 1$. %

\begin{wrapfigure}{r}{0.5\columnwidth}
    \includegraphics[width=0.5\columnwidth]{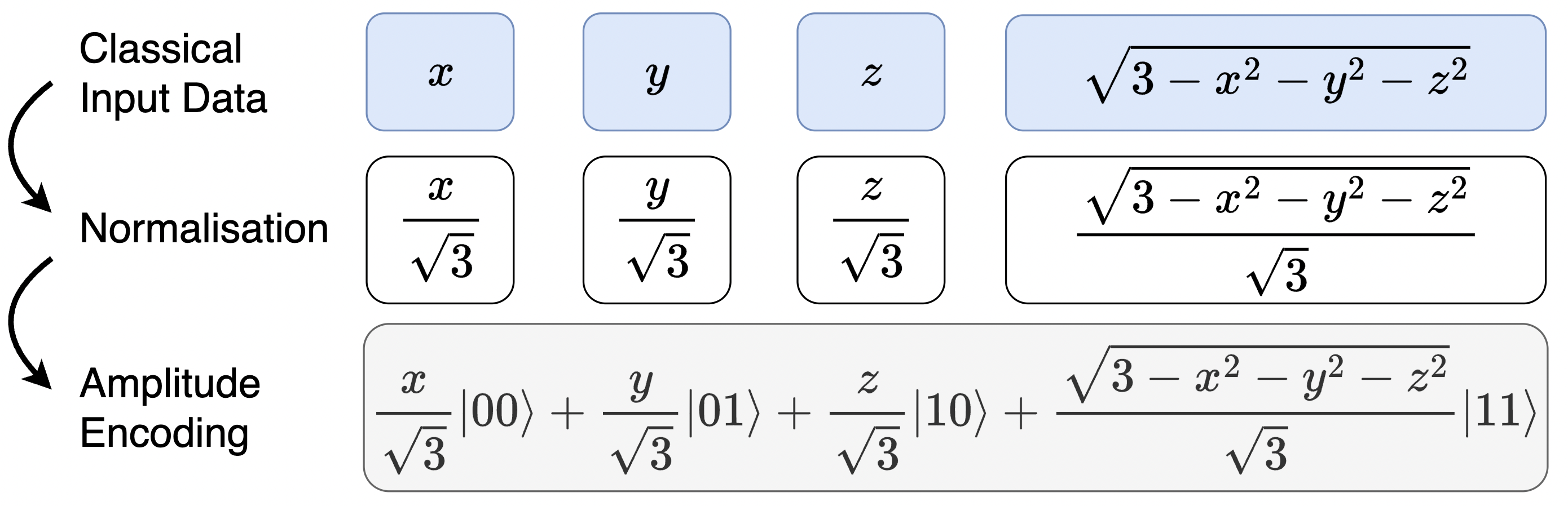} 
    \caption{Data Encoding ($V{=}1$). 
    We encode a 3D vertex by adding an auxiliary value, normalising it to a norm $1.0$ and turning it into a quantum state vector via amplitude encoding. 
    } 
    \label{fig:amplitude_encoding} 
\end{wrapfigure}

\noindent\textbf{Data Normalisation.} 
Since the output of the auto-encoder always lies in the positive octant, we also normalise our data to lie in the positive octant. 
We first compute a tight, cubic bounding box around all vertices in the dataset. 
We then normalise our data by shifting and isotropically rescaling it such that the bounding box coincides with the unit cube in the positive octant. 
See our supplement for further details.

\noindent\textbf{Encoding the Classical Data.} 
Our inputs contain dozens of vertices, which is substantial for current quantum systems. 
We use amplitude encoding to transform our classical data into a quantum state vector, since it allows us to utilise the exponentially large Hilbert space; 
see Fig.~\ref{fig:amplitude_encoding}. 
Recall that amplitude encoding demands 
a state vector with the unit norm. 
Hence, we introduce an auxiliary value $\sqrt{3-x_i^2-y_i^2-z_i^2}$ for each normalised vertex $\mathbf{\tilde{v}_i} = (x_i,y_i,z_i)$ as a constant normalisation factor across all point clouds. 
Moreover, the state vector size is always a power of 2 and 
we fill up the remaining entries with zeros to get a state vector of size $2^N$. 
Finally, the norm of the state vector is one and we, thus, need to normalise all values by $\sqrt{3V}$:
{\small
\begin{equation}
\begin{split}
    \ket{\psi_{\mathit{in}}} = \frac{1}{\sqrt{3V}}\sum_{i=0}^{V-1} \bigg( x_i\ket{3i}+y_i\ket{3i+1}+z_i\ket{3i+2} 
    + \sqrt{3-x_i^2-y_i^2-z_i^2}\ket{3V+i} \bigg) 
    + \sum_{j=4V}^{2^N-1} 0\ket{j}.
\end{split}
\end{equation}
}

\vspace{-9pt}

\subsection{Quantum Circuit Design}\label{sec:designing_circuit} 
\begin{figure}
\centering
    \includegraphics[width=\columnwidth]{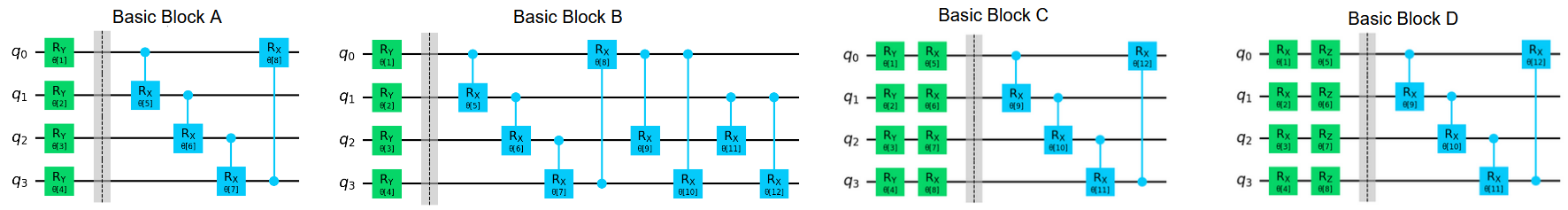}
    \caption{Different quantum blocks. ``A'' and ``D'' are inspired by Sim \textit{et al.}~\cite{Sim_2019}.  %
    }
    \label{fig:all_circuits}
\end{figure}

We next describe the design of our architecture with quantum  circuits. 
The \emph{basic block } of such a quantum circuit is a layer of rotation gates $L_R$ followed by a layer $L_{\mathit{CR}}$ of controlled rotation gates. 
Naturally, there are several possible combinations of these gates.
Thus, we investigate different architectures for these basic blocks (Fig.~\ref{fig:all_circuits}). 
We start with the simple architecture of the basic block ``A'', which uses \emph{linear} entanglement where each qubit is entangled with two neighbours in a circular fashion. 
We experimented with different combinations of gate types and found  $R_Y$ gates in the $L_R$ layer and $CR_{X}$ gates in the $L_{\mathit{CR}}$ layer to work best. %
This is consistent with prior work that measures the expressibility and entangling capacity of circuits under different combinations of these rotation and controlled rotation gates \cite{Sim_2019}.  
The design of ``B'' is a variant of ``A'' that additionally considers the bottleneck in its design. 
Specifically, we add entangling gates between qubits that get removed at the bottleneck and those that remain (as we will discuss later in Sec.~\ref{sec:architecture}), enabling better information propagation. 
Finally, the blocks ``C'' and ``D'' are alternative ways of increasing the capacity of the block ``A'' without adding entangling gates. 

We combine two basic   
blocks (or their inverses), which we call $F$ and $S$, into a \emph{block} $X=S F$. 
We investigate two types: (1) the ``repeat'' type uses the same architecture for both basic blocks, while (2) the ``inverse'' type uses the inverse architecture of $F$ for $S$; see supplement C for a visualisation. %
We initialise the parameters of $F$ randomly. 
For $S$, we investigate two initialisation schemes: (1) the ``random'' scheme initialises it with random parameters, while (2) the ``identity'' scheme depends on the type of the block. 
In the second case, for the ``repeat'' type, we use the same initial parameters as for $F$, and for the ``inverse'' type, we use the initial parameters that make $S$ the inverse of $F$. 
The ``inverse'' type with the ``identity'' initialisation is also known as \emph{identity-block initialisation} \cite{Grant2019initialization} that is meant to prevent the model from being stuck in a barren plateau at the beginning of training. 

Finally, we build a quantum circuit by chaining $J$ blocks together: $U=X_{J-1}\cdots X_0$. 
We always use the same architecture scheme and initialisation scheme for all $X_j$ in a circuit $U$. 

\subsection{Full Architecture}\label{sec:architecture}
We next employ 
any such design 
of a single quantum circuit (Sec.~\ref{sec:designing_circuit}) to build the overall architecture of our quantum auto-encoder, whose parameters are denoted by $\theta$; see Fig.~\ref{fig:full_QA}. 

Specifically, both the encoder and decoder each consist of a single quantum circuit with the same number of blocks $J$, mirroring the classical fully connected design.

\noindent\textbf{Encoder.} 
The encoder reduces the dimension of its input, which is non-trivial to implement when using a quantum circuit because it is a unitary transform. 
To that end, we define the encoder in terms of two subsystems of qubits, namely $A$ with $N_A$ qubits and $B$ with $N_B{=}N{-}N_A$ qubits.  
$A$ contains the information ultimately passed on to the decoder and $B$ is discarded at the end of the encoder. 
The input state vector $\ket{\psi_{\mathit{in}}}_{AB}=\ket{\psi_{\mathit{in}}}$ is the state vector of all qubits in $A$ and $B$ and the encoder $E$ first acts on it as $\ket{\phi}_{AB} = E\ket{\psi_{\mathit{in}}}_{AB}$.

\noindent\textbf{Bottleneck.} 
To map into the latent space, we discard the information present in the qubit subsystem $B$ by \emph{tracing out} its qubits. 
This is analogous to marginalising out $N_B$ binary dimensions of an $N$-dimensional probability distribution if we treat the \emph{squared} amplitudes of $\ket{\phi}_{AB}$ as a distribution. 
This partial trace introduces a quantum non-linearity into the otherwise unitary auto-encoder. 
This trace sums over all basis states $\ket{i}_B\in(\mathbb{C}^{2})^{\otimes N_B}$ of $B$: 
{%
\begin{equation}
    \rho = \operatorname{Tr}_{B}[\ket{\phi}_{AB}] = \sum_i \underbrace{(I_A\otimes \bra{i}_B)}_{2^{N_A} \times 2^N} \underbrace{(\ket{\phi}_{AB}\bra{\phi}_{AB})}_{2^{N} \times 2^N} \underbrace{(I_A\otimes \ket{i}_B)}_{2^{N} \times 2^{N_A}}, 
\end{equation}}
\hspace{-3pt}where $\ket{\phi}_{A} \in (\mathbb{C}^{2})^{\otimes N_A}$, $\bra{\phi} = \ket{\phi}^\dag$ is a Hermitian conjugate and, hence, a row vector, and $I_A$ is the $2^{N_A}{\times}2^{N_A}$ identity. 
We treat $I_A\otimes \bra{i}_B$, which is a $2^{N_A}{\times}2^{N_A}{\times}2^{N_B}$ tensor, as its flattened $2^{N_A}{\times}2^{N_A} 2^{N_B}$ version. 
On its diagonal, the $2^{N_A}{\times}2^{N_A}$ matrix $\rho$ %
contains the squared amplitudes of $\ket{\phi}_{A}$. 
We use amplitude encoding to turn these amplitudes into the state $\ket{\phi}_{A}$, which is the latent code of the auto-encoder, \emph{i.e.}\ a learned quantum 3D point cloud representation.

\noindent\textbf{Decoder.} 
We face an analogous problem with the decoder $D$ as with the encoder: We need to map from a lower-dimensional space to a higher-dimensional one, using a unitary transform. 
We achieve this by expanding $\ket{\phi}_{A}$ with qubits of $B$, which are freshly initialised to $\ket{0}_B$. 
We then obtain $\ket{\xi}=\ket{\xi}_{AB} = D (\ket{\phi}_{A}\otimes \ket{0}_B)$. %

\subsection{Training}\label{sec:training}
To determine the best set of parameters $\theta$, we need to define a loss function and then update the parameters accordingly. 
However, \textit{measuring}  $\ket{\xi}$ yields a single $N$-bit output string, which contains little information and is not differentiable, preventing effective, gradient-based optimisation. 
We circumvent this by running the auto-encoder multiple times. 
This yields an empirical estimate of the frequency of each bit string, \textit{i.e.}~of the probability of each dimension of the state vector;
in simulation, a single run of the auto-encoder is sufficient to obtain all probabilities.
We treat these probability amplitudes $(\alpha_0,\ldots,\alpha_{2^N-1})$ as the output and process them in a classical loss function. 
We treat $(\alpha_0, \alpha_1, \ldots, \alpha_{3V-1})$ as the predicted vertices and $(\alpha_{3V},\ldots,\alpha_{4V-1})$ 
as the predicted auxiliary values. 
We then encourage these output amplitudes of $\ket{\xi}$ to be similar to the input amplitudes of $\ket{\phi_{\mathit{in}}}$ with the loss $\mathcal{L}_{rec}$:  
\begin{equation}
\scriptsize 
\mathcal{L}_{rec} = \sum_{i=0}^{V-1} \Bigg(\sqrt{ \boldsymbol{\zeta_i} } +
\Bigg|\alpha_{3V+i}-\frac{\sqrt{3-x_i^2-y_i^2-z_i^2}}{\sqrt{3V}}\Bigg| \Bigg) + \sum_{j=4V}^{2^N-1} |\alpha_j|, 
\text{with}\;\,\boldsymbol{\zeta_i} = \Big(\alpha_{3i}{-}\frac{x_i}{\sqrt{3V}}\Big)^2+\Big(\alpha_{3i{+}1}{-}\frac{y_i}{\sqrt{3V}}\Big)^2+\Big(\alpha_{3i{+}2}{-}\frac{z_i}{\sqrt{3V}}\Big)^2. 
\end{equation}

To minimise $\mathcal{L}_{rec}$, we use the Adam optimiser \cite{kingma2014adam} for backpropagation with an initial learning rate of $10^{-2}$ and beta values of $0.9$ and $0.99$. 
We apply a loss-based learning rate schedule, use a batch size of $1$, and train for $10k$ epochs. 
We use the Pytorch \cite{NEURIPS2019_9015} interface in PennyLane \cite{Bergholm2018} for noise-free quantum simulation on the CPU.

\section{Experimental Evaluation} 

We next evaluate the performance of our method  qualitatively and quantitatively. 
We use %
the AMASS dataset \cite{AMASS:ICCV:2019} with
3D human motion capture data (joint locations) for various poses with a high variety of articulations. 
We thus quantify the reconstruction error via the mean Euclidean distance in $cm$ across all joints. 
We temporally downsample the data to $12$ fps and select the first $10k$ poses in temporal order. 
The resulting data is split into training and test data with a 80:20 split.
This corresponds to a training motion of $16$ $sec$ and a test motion of $4$ $sec$ in each chunk. 
We use $16$ joints from the SMPL model~\cite{SMPL2015} and normalise out the global transform. 
We encode human poses with $V{=}16$ vertices into state vectors of length $2^6$, corresponding to a six-qubit circuit. 
At the bottleneck, we remove two qubits, which yields a latent code of length $16$. 
Unless stated otherwise, we use $J{=}8$ blocks with block ``B'' in the \textit{repeat} architecture with the \textit{identity} initialisation, as we found these to work best.

\subsection{Comparisons} 

\begin{wrapfigure}{r}{0.5\columnwidth}
\centering
    \includegraphics[width=0.5\columnwidth]{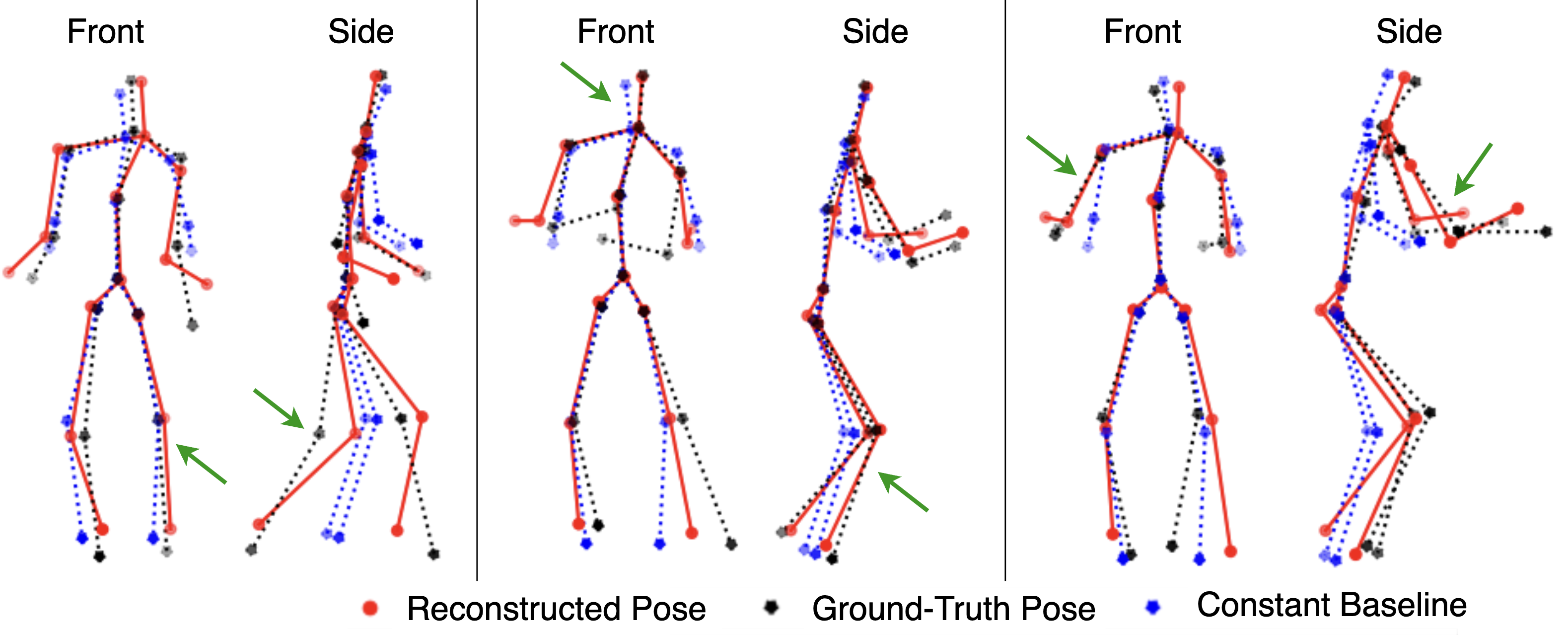}
    \caption{Qualitative results of our fully quantum architecture. 
    }     
    \label{fig:qualitative_results}
\end{wrapfigure}
Fig.~\ref{fig:qualitative_results} shows qualitative results using the proposed method. 
The reconstructed poses follow the ground-truth states well, with some occasional coarse details.

\noindent\textbf{Baselines.} 
We compare to three classical (non-quantum) baselines. 
First, independent of the input, the \emph{constant baseline} always predicts the mean mesh, \textit{i.e.}~the average of all the meshes in the training data. 
Second, as described in Sec.~\ref{sec:intro}, a basic classical architecture using fully connected layers as the encoder and decoder is quite powerful. 
We define such a \emph{classical fully connected baseline} by replacing the quantum encoder and decoder with a fully connected layer each in our architecture. 
We use an ELU non-linearity \cite{clevert2015fast} at the bottleneck and 
match the number of parameters in the fully connected layers with those in the quantum architecture. 
We do not use auxiliaries and do not fill up the input vector with zeros. 
Third, to better locate performance bottlenecks, we also experiment with a \emph{classical mimic baseline} whose design sticks very closely to the quantum architecture. 
The \emph{only} change we make to our architecture is replacing quantum circuits with square fully connected layers whose output we normalise to norm $1$.

\noindent\textbf{Quantitative Results.}
Quantitative results are reported in Tab.~\ref{tab:classical_comparison}.  
We find that 3D-QAE outperforms the constant baseline but is barely competitive with the best classical baseline. %
Note that the fully connected baseline matched to four blocks achieves a mean Euclidean distance of $11.51$ $cm$. %
In addition, the mimic baseline performs close to the fully connected baseline, which indicates that the design choices in which they differ 
are, most probably, not the limiting factor for the performance of our method. 

\begin{table}[H]
\centering
\begin{tabular}{l|c}
\hline
Method & mean~Euclidean distance \\ 
\hline
Constant baseline & $13.58$ \\ 
Classical mimic baseline & $3.75$ \\ 
Fully connected baseline ($8$ blocks) & $8.37$ \\ 
Fully connected baseline ($16$ blocks) & $3.85$ \\ 
\hline
Ours ($8$ blocks) & $10.86$ \\
Ours ($16$ blocks) & $10.45$ \\
\end{tabular}
\caption{Comparisons against classical baselines. We match the number of parameters in the classical fully connected baseline to those of our architecture with $8$ or $16$ blocks.} 
\label{tab:classical_comparison}
\end{table}

\begin{wraptable}{r}{0.55\columnwidth}
\centering
\resizebox{0.55\columnwidth}{!}{
\begin{tabular}{l|l|c|c|c|c}
\hline
\multirow{2}{*}{\textbf{Architecture}} & \multicolumn{1}{c|}{\multirow{2}{*}{\textbf{Initialisation}}} & \multicolumn{4}{c}{\textbf{Basic Block}} \\ \cline{3-6} 
 & & ``A'' & ``B'' & ``C'' & ``D'' \\ \hline
\multirow{2}{*}{Repeat} 
& Random   & 12.86 & 12.58 & 12.16 & 12.52 \\ \cline{2-6} 
& Identity & 11.36 & $\mathbf{10.86}$ &12.03 & 11.52 \\ \hline
\multirow{2}{*}{Inverse} 
& Random   & 13.39 & 11.94 & 12.56 & 11.82 \\ \cline{2-6} 
& Identity & 12.26 & 11.45 & 12.41 & 12.11 \\ 
\end{tabular}
}
\caption{M. Euclidean distance for different circuit designs for $J{=}8$ blocks in the encoder and decoder. 
} 
\label{table:same_blocks}
\end{wraptable}

\subsection{Analysis}\label{sec:analysis}

\textbf{Circuit Design.} 
We investigate the performance along three axes: (1) which basic block to use, (2) \textit{repeat} \textit{vs}\ \textit{inverse} design, and (3) \textit{random} \textit{vs}\ \textit{identity} initialisation. 
As is common in the QML literature, we first compare the different architectures using the same number of blocks, namely $J{=}8$; 
see 
Tab.~\ref{table:same_blocks}. 
We see that basic block ``B'' with the \textit{repeat} design and \textit{identity} initialisation performs the best. 
In general, barren plateaus often arise 
with generic, hardware-efficient circuits like ours. 
However, we find the \textit{identity} initialisation (\emph{i.e.}\ inverse architecture with identity initialisation) that mitigates them yields no advantage in our case. 
Thus, barren plateaus appear to not be a limiting factor for the current best-performing circuit.

\begin{wraptable}{r}{0.55\columnwidth}
\centering
\resizebox{0.52\columnwidth}{!}{
\begin{tabular}{l|l|c|c|c|c}
\hline
\multirow{2}{*}{\textbf{Architecture}} & \multirow{2}{*}{\textbf{Initialisation}} & \multicolumn{4}{c}{\textbf{Basic Block}} \\ \cline{3-6} 
 & & ``A'' & ``B'' & ``C'' & ``D'' \\ \hline
\multirow{2}{*}{Repeat}   
& Random   & $11.52$ & $12.58$ & $11.73$ & $11.12$ \\ \cline{2-6} 
& Identity & $11.37$ & $\mathbf{10.86}$ & $11.74$ & $11.95$ \\ \hline
\multirow{2}{*}{Inverse} 
& Random   & $11.99$ & $11.94$ & $12.34$ & $12.33$ \\ \cline{2-6} 
& Identity & $12.38$ & $11.45$ & $11.87$ & $11.92$ \\ 
\end{tabular}
}
\caption{M. Euclidean distance for different circuit designs. We use $J{=}8$ blocks for basic block ``B''. For the other basic blocks, we use that $J$ that matches the number of parameters of the basic block ``B'' architecture the closest.} 
\label{table:same_parameters} 
\end{wraptable}
However, the number of quantum gates (and, in turn, parameters) differs across the different block types. 
We, thus, compare the different block types by matching the number of parameters to those of ``B''. 
To this end, we increase the number of blocks for the ``A'', ``C'', and ``D'' types accordingly. 
The results are in Tab.~\ref{table:same_parameters}. 
While those circuits using basic blocks other than ``B'' improve their performance, ``B'' still 
shows the best performance overall. 
This validates our choice to add entangling gates between qubits that were removed and qubits that remained at the bottleneck.

\begin{figure}
\centering
    \includegraphics[width=0.7\columnwidth, keepaspectratio]{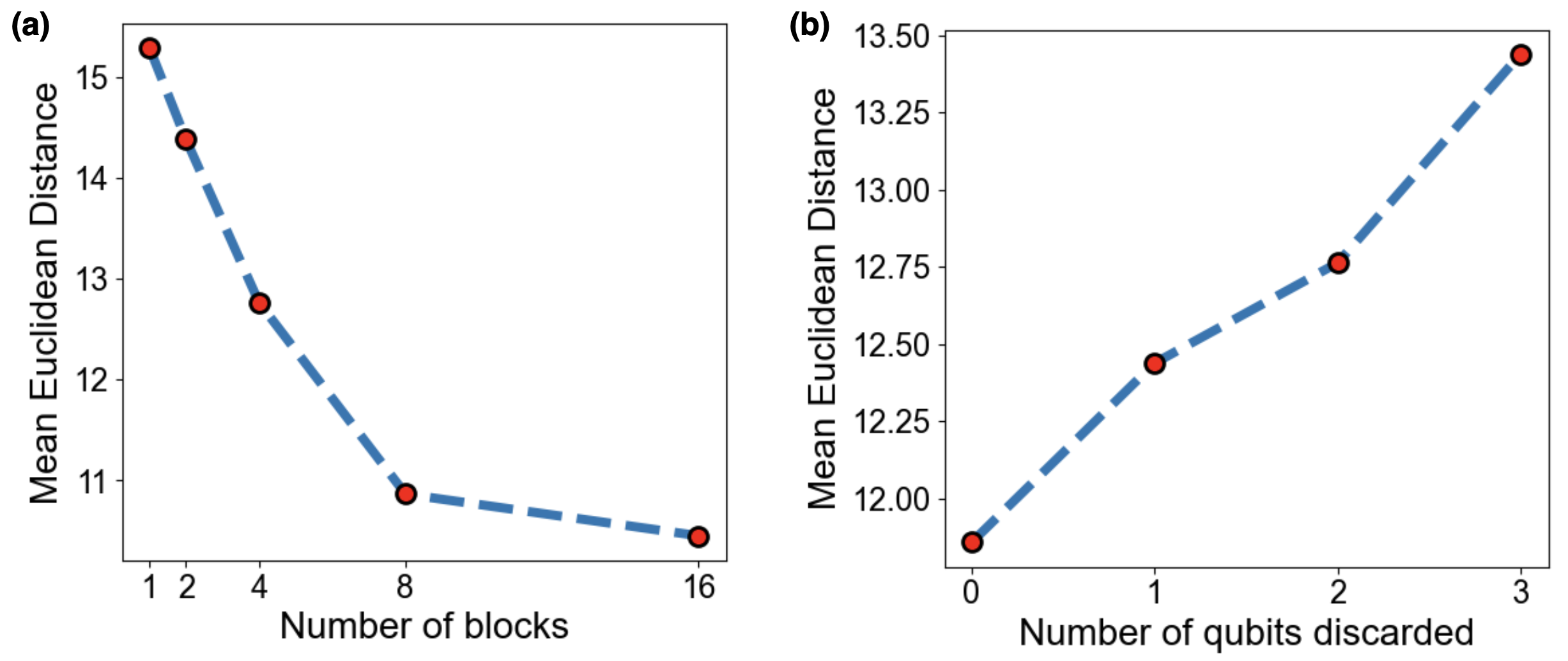}
    \caption{Mean Euclidean distance for a varying number of (a) blocks and (b) qubits discarded in the bottleneck.
    } 
    \label{fig:graphs}
\end{figure} 

\begin{figure}
    \begin{tabular}{cc}
    \bmvaHangBox{\fbox{\includegraphics[height=0.15\textheight,keepaspectratio]{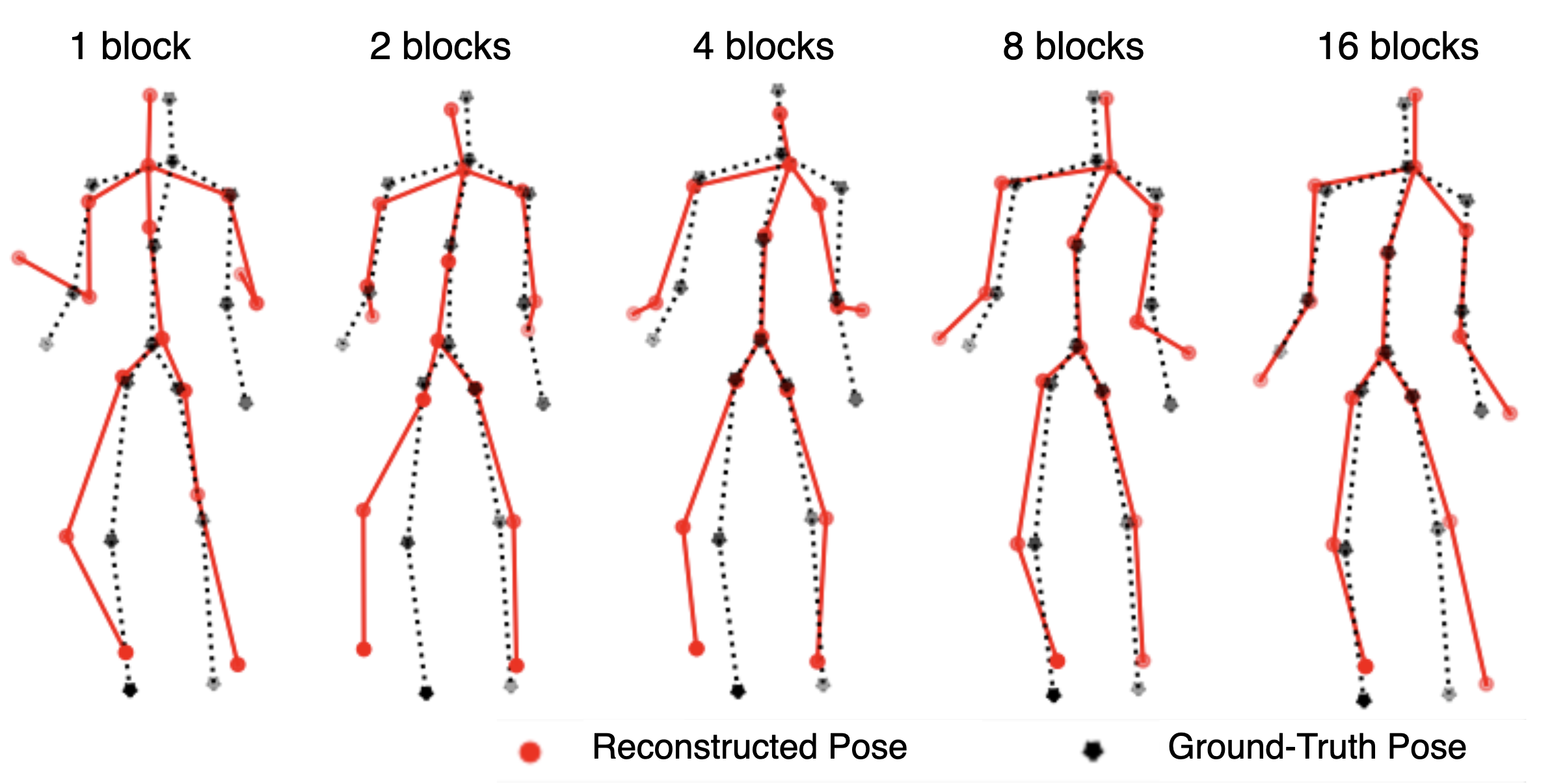}}}&
    \bmvaHangBox{\fbox{\includegraphics[height=0.15\textheight,keepaspectratio]{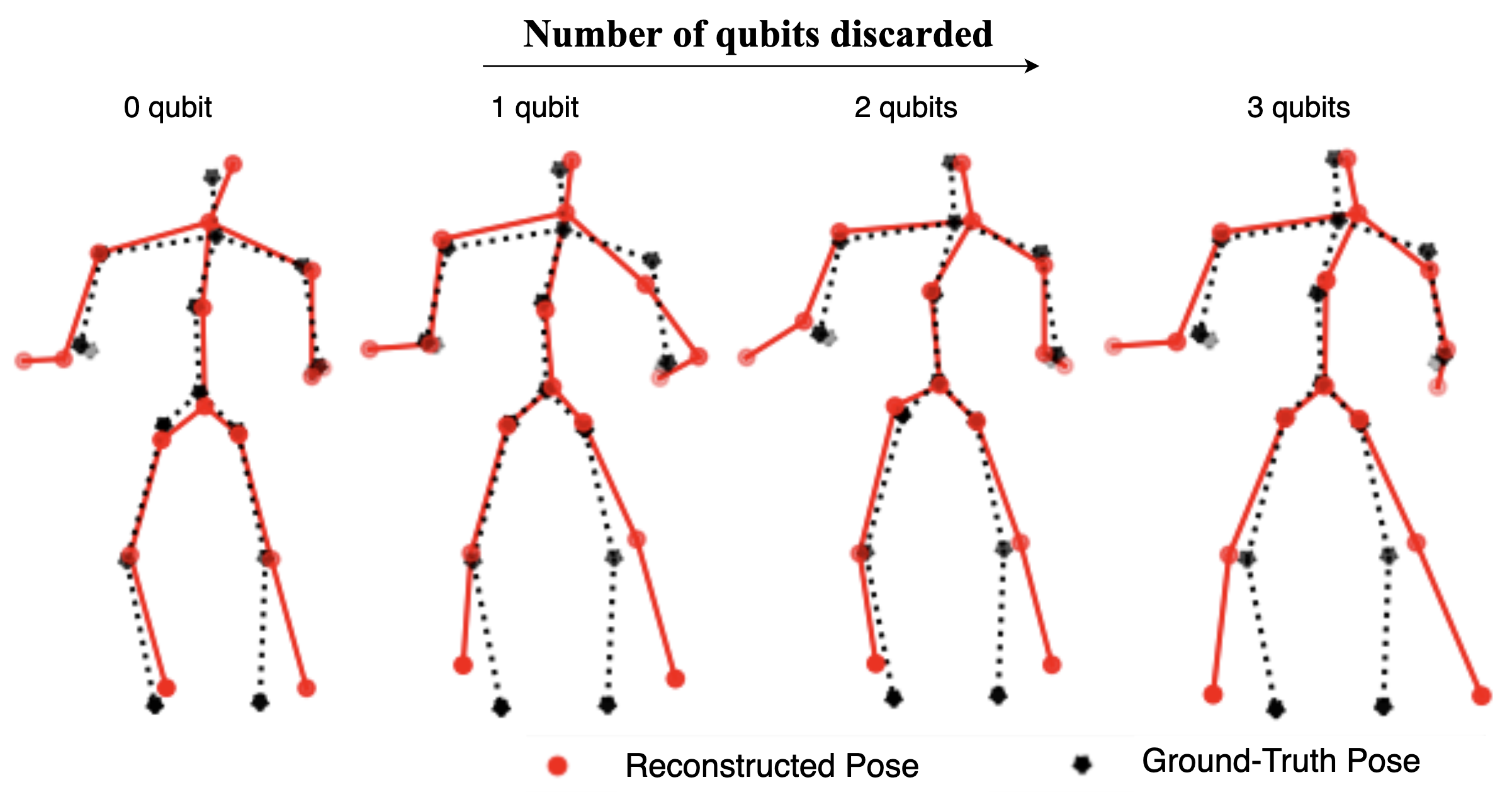}}}\\
    (a)&(b)
    \end{tabular}
    \caption{Examples for a varying numbers of (a) blocks $J$ and (b) bottelneck qubits.
    }
    \label{fig:increasing_blocks}
\end{figure}

\noindent\textbf{Number of Blocks.} 
Like classical neural architectures, quantum circuits become more expressive (and harder to implement and optimise) the deeper they are. 
Here, we take a closer look at how the quality and the quantitative error change with the number of blocks $J$. 
Fig.~\ref{fig:increasing_blocks}(a) contains qualitative results that show a clear improvement with more blocks. 
Fig.~\ref{fig:graphs}(a) confirms this quantitatively. 
Using $J{=}16$ rather than $8$ blocks improves results somewhat. However, this comes at an impracticable training time, as it doubles the $45$ hours it takes on the CPU for $J{=}8$. 

\noindent\textbf{Number of Bottleneck Qubits.} 
We discard 
several qubits at the bottleneck 
and thereby reduce 
the amount of information transmitted into the decoder. 
Figs.~\ref{fig:graphs}(b) and \ref{fig:increasing_blocks}(b) show numerical and qualitative results, respectively. 
As expected, a smaller latent space leads to worse reconstruction accuracy. 
We note that, in contrast to the classical setting, every discarded qubit reduces the amount of information by a factor of two instead of linearly. %

\begin{wraptable}{r}{0.55\columnwidth}
\centering
\resizebox{0.52\columnwidth}{!}{
\begin{tabular}{c|c||cc||c}
\hline
Number of Blocks & ELU & QE-CD & CE-QD & Ours \\ \hline
$4$ & $12.97$ & $7.06$ & $11.25$ & $12.76$\\ 
$8$ & $10.98$ & $6.34$ & $10.20$ & $10.86$\\                
\end{tabular}
}
\caption{Mean Euclidean distances. Instead of marginalisation, \emph{ELU} takes a sub-vector from the encoder and applies an ELU at the bottleneck. \emph{QE-CD} is a hybrid with a quantum encoder and a classical decoder, \emph{CE-QD} uses a classical encoder and a quantum decoder. %
}
\label{tab:more_quantitative}
\end{wraptable}

\noindent\textbf{Non-Linearity.} 
Tab.~\ref{tab:more_quantitative} shows that 3D-QAE with a quantum non-linearity (QNL) is on par with a carefully designed classical activation (ELU~\cite{clevert2015fast}). 
This is supported by the mimic baseline (with QNL) that is on par with the fully connected baseline (with ELU); see Tab.~\ref{tab:classical_comparison} for numerical comparisons. 

\noindent\textbf{Hybrid Models.}
We next shortly look at hybrid models. 
We replace either only the encoder or only the decoder with a classical fully connected layer as in the mimic baseline. 
Tab.~\ref{tab:more_quantitative} shows that replacing quantum parts with classical ones improves the performance. 

\noindent\textbf{Simpler Tasks.} 
Finally, we explore simpler auto-encoding tasks to better understand the gap between the constant baseline and our method. 
To this end, we look at three different body parts with three vertices each, rather than the entire body. 
Tab.~\ref{tab:body_parts} shows that our method outperforms the constant baseline on these easier tasks even more than on the full body. 
This suggests that task difficulty (\textit{e.g.,} motion complexity) substantially influences the performance of our method. The training time decreases exponentially with the number of qubits. For these simpler experiments in the $J=8$ case, it requires 9 hours on the CPU.

\begin{table}[H]
\centering
    \begin{tabular}{l|ccc|c}
    \hline
             Method       & Left Leg & Right Arm & Spine & Full Body\\
                    \hline
      Constant Baseline   &  $10.7$ & $22.1$ & $7.4$ & $13.6$\\
      Ours (8 blocks)          &  $\mathbf{3.2}$ & $\mathbf{6.5}$ & $\mathbf{1.5}$ & $\mathbf{10.9}$\\
      \hline
      Ratio (Ours / Con.)  & $0.30$ & $0.29$ &$0.20$&  $0.80$
    \end{tabular}
    \caption{Auto-encoding body parts. Mean Euclidean distances are in $cm$.} 
    \label{tab:body_parts}
\end{table}

\subsection{Discussion}
Since quantum supremacy has not yet been achieved in practical settings, it was to be expected that our method would not outperform classical baselines. 
We have eliminated a number of possible reasons for the gap between our best-performing fully quantum architecture and the classical baselines. 
The performance of the classical mimic baseline suggests that the limiting factor lies with the quantum circuit itself. 
However, as discussed, barren plateaus appear to not be an issue, as mitigating them has no discernible effect on performance. 
Furthermore, basic block ``B'' designed with the bottleneck in mind performs slightly better than the alternatives. 
Still, we found only small performance differences over a wide range of typical hardware-efficient circuit designs. 
The hybrid models indicate that enough information remains at the bottleneck when using a quantum encoder. 
The issue thus rather lies with the expressiveness of the quantum decoder (perhaps because it is unitary and, thus, represents an \emph{orthonormal} set of basis shapes for the output). 
However, while further increasing the number of blocks does improve the accuracy, the returns diminish quickly. 
Even larger architectures are infeasible in reasonable runtime with current software and hardware.

\noindent\textbf{Future Work.} Many quantum papers \cite{Shi2020TrainingAQ, https://doi.org/10.48550/arxiv.2208.04988, https://doi.org/10.48550/arxiv.2109.01831, Landman_2022,Yang_2022_CVPR} that investigate vision tasks focus on classification rather than regression  (\emph{e.g.}\ digit classification on  MNIST~\cite{deng2012mnist}). 
Future work on applying quantum computing to 3D tasks could thus also address classification problems. 
In a narrower sense, work on equivariant quantum networks \cite{https://doi.org/10.48550/arxiv.2210.08566, https://doi.org/10.48550/arxiv.2210.07980, https://doi.org/10.48550/arxiv.2210.09974} that explicitly accounts for global translation and rotation are an interesting future avenue. %

\section{Conclusion}
This paper investigates 3D point cloud auto-encoding with a fully quantum architecture. 
Our data normalisation scheme with amplitude encoding is crucial and allows representing classical data (3D point clouds) as inputs for our approach. 
We explore a wide range of design choices and succeed in overcoming a surprisingly challenging hurdle, \textit{i.e.}~outperforming a constant baseline. 
We observe in the experiments that replacing either the encoder or decoder with their classical counterparts decreases the reconstruction error. 
Moreover, the simpler the motions, the lower the error our fully quantum method achieves. 
Our results suggest that a substantial gap remains between classical and quantum architectures, which is in line with prior work in QML. 
This first study and our thorough analysis plausibly eliminate many reasons for this gap, and future work could further attempt to reduce and eventually close it. 

\noindent\textbf{Acknowledgement.} 
This work was supported by the ERC consolidator grant \textit{4DReply} (770784).

\bibliography{egbib}

\begin{thebibliography}{76}
\providecommand{\natexlab}[1]{#1}
\providecommand{\url}[1]{\texttt{#1}}
\expandafter\ifx\csname urlstyle\endcsname\relax
  \providecommand{\doi}[1]{doi: #1}\else
  \providecommand{\doi}{doi: \begingroup \urlstyle{rm}\Url}\fi

\bibitem[Abbas et~al.(2021)Abbas, Sutter, Zoufal, Lucchi, Figalli, and
  Woerner]{Abbas_2021}
Amira Abbas, David Sutter, Christa Zoufal, Aurelien Lucchi, Alessio Figalli,
  and Stefan Woerner.
\newblock The power of quantum neural networks.
\newblock \emph{Nature Computational Science}, 2021.

\bibitem[{Achache} et~al.(2021){Achache}, {Horesh}, and
  {Smolin}]{2020arXiv201214714A}
Tom {Achache}, Lior {Horesh}, and John {Smolin}.
\newblock {Denoising quantum states with Quantum Autoencoders -- Theory and
  Applications}.
\newblock \emph{Conference on Quantum Information Processing (QIP)}, 2021.

\bibitem[Alam and Ghosh(2022)]{9885859}
Mahabubul Alam and Swaroop Ghosh.
\newblock Deepqmlp: A scalable quantum-classical hybrid deep neural network
  architecture for classification.
\newblock In \emph{International Conference on VLSI Design and International
  Conference on Embedded Systems (VLSID)}, 2022.

\bibitem[Alam et~al.(2021)Alam, Kundu, Topaloglu, and Ghosh]{alam2021quantum}
Mahabubul Alam, Satwik Kundu, Rasit~Onur Topaloglu, and Swaroop Ghosh.
\newblock Quantum-classical hybrid machine learning for image classification.
\newblock In \emph{IEEE/ACM International Conference On Computer Aided Design
  (ICCAD)}, 2021.

\bibitem[Amin et~al.(2018)Amin, Andriyash, Rolfe, Kulchytskyy, and
  Melko]{Amin_2018}
Mohammad~H. Amin, Evgeny Andriyash, Jason Rolfe, Bohdan Kulchytskyy, and Roger
  Melko.
\newblock Quantum boltzmann machine.
\newblock \emph{Physical Review X}, 2018.

\bibitem[Bagautdinov et~al.(2018)Bagautdinov, Wu, Saragih, Fua, and
  Sheikh]{8578506}
Timur Bagautdinov, Chenglei Wu, Jason Saragih, Pascal Fua, and Yaser Sheikh.
\newblock Modeling facial geometry using compositional vaes.
\newblock In \emph{Computer Vision and Pattern Recognition (CVPR)}, 2018.

\bibitem[Bergholm et~al.(2018)]{Bergholm2018}
Ville Bergholm et~al.
\newblock Pennylane: Automatic differentiation of hybrid quantum-classical
  computations.
\newblock \emph{arXiv e-prints}, 2018.

\bibitem[Bhatia et~al.(2023)Bhatia, Tretschk, Lähner, Benkner, Moeller,
  Theobalt, and Golyanik]{Bhatia2023CCuantuMM}
Harshil Bhatia, Edith Tretschk, Zorah Lähner, Marcel Benkner, Michael Moeller,
  Christian Theobalt, and Vladislav Golyanik.
\newblock Ccuantumm: Cycle-consistent quantum-hybrid matching of multiple
  shapes.
\newblock In \emph{Computer Vision and Pattern Recognition (CVPR)}, 2023.

\bibitem[Biamonte et~al.(2017)Biamonte, Wittek, Pancotti, Rebentrost, Wiebe,
  and Lloyd]{Biamonte_2017}
Jacob Biamonte, Peter Wittek, Nicola Pancotti, Patrick Rebentrost, Nathan
  Wiebe, and Seth Lloyd.
\newblock Quantum machine learning.
\newblock \emph{Nature}, 2017.

\bibitem[{Birdal} et~al.(2021){Birdal}, {Golyanik}, {Theobalt}, and
  {Guibas}]{QuantumSync2021}
Tolga {Birdal}, Vladislav {Golyanik}, Christian {Theobalt}, and Leonidas
  {Guibas}.
\newblock Quantum permutation synchronization.
\newblock In \emph{Computer Vision and Pattern Recognition (CVPR)}, 2021.

\bibitem[Bokhan et~al.(2022)Bokhan, Mastiukova, Boev, Trubnikov, and
  Fedorov]{Bokhan2022}
Denis Bokhan, Alena~S. Mastiukova, Aleksey~S. Boev, Dmitrii~N. Trubnikov, and
  Aleksey~K. Fedorov.
\newblock Multiclass classification using quantum convolutional neural networks
  with hybrid quantum-classical learning.
\newblock \emph{Front. in Phys.}, 10:\penalty0 1069985, 2022.

\bibitem[Bouritsas et~al.(2019)Bouritsas, Bokhnyak, Ploumpis, Bronstein, and
  Zafeiriou]{bouritsas2019neural}
Giorgos Bouritsas, Sergiy Bokhnyak, Stylianos Ploumpis, Michael Bronstein, and
  Stefanos Zafeiriou.
\newblock Neural 3d morphable models: Spiral convolutional networks for 3d
  shape representation learning and generation.
\newblock In \emph{International Conference on Computer Vision (ICCV)}, 2019.

\bibitem[Bourlard and Kamp(1988)]{article}
Herve Bourlard and Y~Kamp.
\newblock Auto-association by multilayer perceptrons and singular value
  decomposition.
\newblock \emph{Biological cybernetics}, 1988.

\bibitem[Bravo-Prieto(2021)]{Bravo_Prieto_2021}
Carlos Bravo-Prieto.
\newblock Quantum autoencoders with enhanced data encoding.
\newblock \emph{Machine Learning: Science and Technology}, 2021.

\bibitem[{Chalumuri} et~al.(2021){Chalumuri}, {Kune}, {Kannan}, and
  {Manoj}]{2021QuIP...20..381C}
Avinash {Chalumuri}, Raghavendra {Kune}, S.~{Kannan}, and B.~S. {Manoj}.
\newblock {Quantum-enhanced deep neural network architecture for image scene
  classification}.
\newblock \emph{Quantum Information Processing}, 2021.

\bibitem[Clevert et~al.(2016)Clevert, Unterthiner, and
  Hochreiter]{clevert2015fast}
Djork-Arn{\'e} Clevert, Thomas Unterthiner, and Sepp Hochreiter.
\newblock Fast and accurate deep network learning by exponential linear units
  (elus).
\newblock \emph{International Conference on Learning Representations (ICLR)},
  2016.

\bibitem[Defferrard et~al.(2016)Defferrard, Bresson, and
  Vandergheynst]{defferrard2016convolutional}
Micha{\"e}l Defferrard, Xavier Bresson, and Pierre Vandergheynst.
\newblock Convolutional neural networks on graphs with fast localized spectral
  filtering.
\newblock \emph{Advances in Neural Information Processing Systems (NeurIPS)},
  2016.

\bibitem[Deng(2012)]{deng2012mnist}
Li~Deng.
\newblock The mnist database of handwritten digit images for machine learning
  research.
\newblock \emph{IEEE Signal Processing Magazine}, 2012.

\bibitem[Doan et~al.(2022)Doan, Sasdelli, Suter, and Chin]{9879497}
Anh-Dzung Doan, Michele Sasdelli, David Suter, and Tat-Jun Chin.
\newblock A hybrid quantum-classical algorithm for robust fitting.
\newblock In \emph{Computer Vision and Pattern Recognition (CVPR)}, 2022.

\bibitem[Elbaz et~al.(2017)Elbaz, Avraham, and Fischer]{Elbaz2017}
Gil Elbaz, Tamar Avraham, and Anath Fischer.
\newblock 3d point cloud registration for localization using a deep neural
  network auto-encoder.
\newblock In \emph{Computer Vision and Pattern Recognition (CVPR)}, 2017.

\bibitem[Farina et~al.(2023)Farina, Magri, Menapace, Ricci, Golyanik, and
  Arrigoni]{Farina2023quantum}
Matteo Farina, Luca Magri, Willi Menapace, Elisa Ricci, Vladislav Golyanik, and
  Federica Arrigoni.
\newblock Quantum multi-model fitting.
\newblock In \emph{Computer Vision and Pattern Recognition (CVPR)}, 2023.

\bibitem[Golyanik and Theobalt(2020)]{golyanik2020quantum}
Vladislav Golyanik and Christian Theobalt.
\newblock A quantum computational approach to correspondence problems on point
  sets.
\newblock In \emph{Computer Vision and Pattern Recognition (CVPR)}, 2020.

\bibitem[Grant et~al.(2019)Grant, Wossnig, Ostaszewski, and
  Benedetti]{Grant2019initialization}
Edward Grant, Leonard Wossnig, Mateusz Ostaszewski, and Marcello Benedetti.
\newblock An initialization strategy for addressing barren plateaus in
  parametrized quantum circuits.
\newblock \emph{{Quantum}}, 2019.

\bibitem[Guijo et~al.(2022)Guijo, Onofre, Del~Bimbo, Mugel, Estepa, De~Carlos,
  Adell, Lojo, Bilbao, and Orus]{https://doi.org/10.48550/arxiv.2208.04988}
Daniel Guijo, Victor Onofre, Gianni Del~Bimbo, Samuel Mugel, Daniel Estepa,
  Xabier De~Carlos, Ana Adell, Aizea Lojo, Josu Bilbao, and Roman Orus.
\newblock Quantum artificial vision for defect detection in manufacturing.
\newblock \emph{arXiv e-prints}, 2022.

\bibitem[Hahner and Garcke(2022)]{hahner2022mesh}
Sara Hahner and Jochen Garcke.
\newblock Mesh convolutional autoencoder for semi-regular meshes of different
  sizes.
\newblock In \emph{Winter Conference on Applications of Computer Vision
  (WACV)}, 2022.

\bibitem[Havl{\'{\i}}{\v{c}}ek et~al.(2019)Havl{\'{\i}}{\v{c}}ek,
  C{\'{o}}rcoles, Temme, Harrow, Kandala, Chow, and Gambetta]{Havl_ek_2019}
Vojt{\v{e}}ch Havl{\'{\i}}{\v{c}}ek, Antonio~D. C{\'{o}}rcoles, Kristan Temme,
  Aram~W. Harrow, Abhinav Kandala, Jerry~M. Chow, and Jay~M. Gambetta.
\newblock Supervised learning with quantum-enhanced feature spaces.
\newblock \emph{Nature}, 2019.

\bibitem[Hu and Ni(2019)]{8895167}
Ling Hu and Qiang Ni.
\newblock Quantum automated object detection algorithm.
\newblock In \emph{International Conference on Automation and Computing
  (ICAC)}, 2019.

\bibitem[Huang et~al.(2021)Huang, Kueng, and Preskill]{Huang2021}
Hsin-Yuan Huang, Richard Kueng, and John Preskill.
\newblock Information-theoretic bounds on quantum advantage in machine
  learning.
\newblock \emph{Phys. Rev. Lett.}, 126:\penalty0 190505, 2021.

\bibitem[Jerbi et~al.(2023)Jerbi, Fiderer, Nautrup, Kübler, Briegel, and
  Dunjko]{Jerbi_2023}
Sofiene Jerbi, Lukas~J. Fiderer, Hendrik~Poulsen Nautrup, Jonas~M. Kübler,
  Hans~J. Briegel, and Vedran Dunjko.
\newblock Quantum machine learning beyond kernel methods.
\newblock \emph{Nature Communications}, 2023.

\bibitem[Kingma and Ba(2015)]{kingma2014adam}
Diederik~P Kingma and Jimmy Ba.
\newblock Adam: A method for stochastic optimization.
\newblock \emph{International Conference on Learning Representations (ICLR)},
  2015.

\bibitem[Kramer(1991)]{Kramer1991NonlinearPC}
Mark~A. Kramer.
\newblock Nonlinear principal component analysis using autoassociative neural
  networks.
\newblock \emph{Aiche Journal}, 1991.

\bibitem[Kwak et~al.(2021)Kwak, Yun, Jung, and Kim]{kwak2021quantum}
Yunseok Kwak, Won~Joon Yun, Soyi Jung, and Joongheon Kim.
\newblock Quantum neural networks: Concepts, applications, and challenges.
\newblock In \emph{International Conference on Ubiquitous and Future Networks
  (ICUFN)}, 2021.

\bibitem[Landman et~al.(2022)Landman, Mathur, Li, Strahm, Kazdaghli, Prakash,
  and Kerenidis]{Landman_2022}
Jonas Landman, Natansh Mathur, Yun~Yvonna Li, Martin Strahm, Skander Kazdaghli,
  Anupam Prakash, and Iordanis Kerenidis.
\newblock Quantum methods for neural networks and application to medical image
  classification.
\newblock \emph{Quantum}, 2022.

\bibitem[Larasati et~al.(2022)Larasati, Le, and Kim]{9932103}
Harashta~Tatimma Larasati, Thi-Thu-Huong Le, and Howon Kim.
\newblock Trends of quantum computing applications to computer vision.
\newblock In \emph{International Conference on Platform Technology and Service
  (PlatCon)}, 2022.

\bibitem[Lau et~al.(2022)Lau, Emani, Chapman, Yao, Lam, Merrill, Warrell,
  Gerstein, and Lam]{Lau2022}
Bayo Lau, Prashant~S Emani, Jackson Chapman, Lijing Yao, Tarsus Lam, Paul
  Merrill, Jonathan Warrell, Mark~B Gerstein, and Hugo Y~K Lam.
\newblock Insights from incorporating quantum computing into drug design
  workflows.
\newblock \emph{Bioinformatics}, 39\penalty0 (1), 12 2022.

\bibitem[Li and Ghosh(2022)]{li2022scalable}
Junde Li and Swaroop Ghosh.
\newblock Scalable variational quantum circuits for autoencoder-based drug
  discovery.
\newblock In \emph{Design, Automation \& Test in Europe Conference \&
  Exhibition (DATE)}, 2022.

\bibitem[Lloyd et~al.(2014)Lloyd, Mohseni, and Rebentrost]{Lloyd_2014}
Seth Lloyd, Masoud Mohseni, and Patrick Rebentrost.
\newblock Quantum principal component analysis.
\newblock \emph{Nature Physics}, 2014.

\bibitem[Locher et~al.(2023)Locher, Cardarelli, and
  M{\"u}ller]{locher2023quantum}
David~F Locher, Lorenzo Cardarelli, and Markus M{\"u}ller.
\newblock Quantum error correction with quantum autoencoders.
\newblock \emph{Quantum}, 2023.

\bibitem[Lockwood and Si(2020)]{Lockwood2021}
Owen Lockwood and Mei Si.
\newblock Playing atari with hybrid quantum-classical reinforcement learning.
\newblock In \emph{NeurIPS Workshop on Pre-registration in Machine Learning},
  2020.

\bibitem[Loper et~al.(2015)Loper, Mahmood, Romero, Pons-Moll, and
  Black]{SMPL2015}
Matthew Loper, Naureen Mahmood, Javier Romero, Gerard Pons-Moll, and Michael~J.
  Black.
\newblock {SMPL}: A skinned multi-person linear model.
\newblock \emph{{ACM} Transactions on Graphics (Proceedings of {SIGGRAPH}
  Asia)}, 2015.

\bibitem[Mahmood et~al.(2019)Mahmood, Ghorbani, Troje, Pons-Moll, and
  Black]{AMASS:ICCV:2019}
Naureen Mahmood, Nima Ghorbani, Nikolaus~F. Troje, Gerard Pons-Moll, and
  Michael~J. Black.
\newblock {AMASS}: Archive of motion capture as surface shapes.
\newblock In \emph{International Conference on Computer Vision (ICCV)}, 2019.

\bibitem[Mangini et~al.(2022)Mangini, Marruzzo, Piantanida, Gerace, Bajoni, and
  Macchiavello]{Mangini_2022}
Stefano Mangini, Alessia Marruzzo, Marco Piantanida, Dario Gerace, Daniele
  Bajoni, and Chiara Macchiavello.
\newblock Quantum neural network autoencoder and classifier applied to an
  industrial case study.
\newblock \emph{Quantum Machine Intelligence}, 2022.

\bibitem[Mari et~al.(2020)Mari, Bromley, Izaac, Schuld, and
  Killoran]{Mari_2020}
Andrea Mari, Thomas~R. Bromley, Josh Izaac, Maria Schuld, and Nathan Killoran.
\newblock Transfer learning in hybrid classical-quantum neural networks.
\newblock \emph{Quantum}, 2020.

\bibitem[Mathur et~al.(2021)Mathur, Landman, Li, Strahm, Kazdaghli, Prakash,
  and Kerenidis]{https://doi.org/10.48550/arxiv.2109.01831}
Natansh Mathur, Jonas Landman, Yun~Yvonna Li, Martin Strahm, Skander Kazdaghli,
  Anupam Prakash, and Iordanis Kerenidis.
\newblock Medical image classification via quantum neural networks.
\newblock \emph{arXiv e-prints}, 2021.

\bibitem[McClean et~al.(2018)McClean, Boixo, Smelyanskiy, Babbush, and
  Neven]{McClean_2018}
Jarrod~R. McClean, Sergio Boixo, Vadim~N. Smelyanskiy, Ryan Babbush, and
  Hartmut Neven.
\newblock Barren plateaus in quantum neural network training landscapes.
\newblock \emph{Nature Communications}, 2018.

\bibitem[Meli et~al.(2022)Meli, Mannel, and Lellmann]{Meli_2022_CVPR}
Natacha~Kuete Meli, Florian Mannel, and Jan Lellmann.
\newblock An iterative quantum approach for transformation estimation from
  point sets.
\newblock In \emph{Computer Vision and Pattern Recognition (CVPR)}, 2022.

\bibitem[Nguyen et~al.(2022)Nguyen, Schatzki, Braccia, Ragone, Coles, Sauvage,
  Larocca, and Cerezo]{https://doi.org/10.48550/arxiv.2210.08566}
Quynh~T. Nguyen, Louis Schatzki, Paolo Braccia, Michael Ragone, Patrick~J.
  Coles, Frederic Sauvage, Martin Larocca, and M.~Cerezo.
\newblock Theory for equivariant quantum neural networks.
\newblock \emph{arXiv e-prints}, 2022.

\bibitem[Nielsen and Chuang(2000)]{NielsenChuang2000}
Michael~A. Nielsen and Isaac~L. Chuang.
\newblock \emph{Quantum Computation and Quantum Information}.
\newblock 2000.

\bibitem[Park et~al.(2021)Park, Dang, Lee, Han, and Moon]{electronics10192406}
Yesul Park, L.~Minh Dang, Sujin Lee, Dongil Han, and Hyeonjoon Moon.
\newblock Multiple object tracking in deep learning approaches: A survey.
\newblock \emph{Electronics}, 2021.

\bibitem[Paszke et~al.(2019)]{NEURIPS2019_9015}
Adam Paszke et~al.
\newblock Pytorch: An imperative style, high-performance deep learning library.
\newblock In \emph{Advances in Neural Information Processing Systems
  (NeurIPS)}. 2019.

\bibitem[Ragone et~al.(2022)Ragone, Braccia, Nguyen, Schatzki, Coles, Sauvage,
  Larocca, and Cerezo]{https://doi.org/10.48550/arxiv.2210.07980}
Michael Ragone, Paolo Braccia, Quynh~T. Nguyen, Louis Schatzki, Patrick~J.
  Coles, Frederic Sauvage, Martin Larocca, and M.~Cerezo.
\newblock Representation theory for geometric quantum machine learning.
\newblock \emph{arXiv e-prints}, 2022.

\bibitem[Ranjan et~al.(2018)Ranjan, Bolkart, Sanyal, and
  Black]{ranjan2018generating}
Anurag Ranjan, Timo Bolkart, Soubhik Sanyal, and Michael~J Black.
\newblock Generating 3d faces using convolutional mesh autoencoders.
\newblock In \emph{European Conference on Computer Vision (ECCV)}, 2018.

\bibitem[Romero et~al.(2017)Romero, Olson, and Aspuru-Guzik]{Romero_2017}
Jonathan Romero, Jonathan~P Olson, and Alan Aspuru-Guzik.
\newblock Quantum autoencoders for efficient compression of quantum data.
\newblock \emph{Quantum Science and Technology}, 2017.

\bibitem[Rumelhart and McClelland(1987)]{6302929}
David~E. Rumelhart and James~L. McClelland.
\newblock Learning internal representations by error propagation.
\newblock In \emph{Parallel Distributed Processing: Explorations in the
  Microstructure of Cognition: Foundations}, 1987.

\bibitem[Sakhnenko et~al.(2022)Sakhnenko, O’Meara, Ghosh, Mendl, Cortiana,
  and Bernab{\'e}-Moreno]{Sakhnenko:2021jme}
Alona Sakhnenko, Corey O’Meara, Kumar~JB Ghosh, Christian~B Mendl, Giorgio
  Cortiana, and Juan Bernab{\'e}-Moreno.
\newblock Hybrid classical-quantum autoencoder for anomaly detection.
\newblock \emph{Quantum Machine Intelligence}, 2022.

\bibitem[Schatzki et~al.(2022)Schatzki, Larocca, Nguyen, Sauvage, and
  Cerezo]{https://doi.org/10.48550/arxiv.2210.09974}
Louis Schatzki, Martin Larocca, Quynh~T. Nguyen, Frederic Sauvage, and
  M.~Cerezo.
\newblock Theoretical guarantees for permutation-equivariant quantum neural
  networks.
\newblock \emph{arXiv e-prints}, 2022.

\bibitem[Sebastianelli et~al.(2021)Sebastianelli, Zaidenberg, Spiller, Le~Saux,
  and Ullo]{sebastianelli2021circuitbased}
Alessandro Sebastianelli, Daniela~Alessandra Zaidenberg, Dario Spiller,
  Bertrand Le~Saux, and Silvia~Liberata Ullo.
\newblock On circuit-based hybrid quantum neural networks for remote sensing
  imagery classification.
\newblock \emph{Journal of Selected Topics in Applied Earth Observations and
  Remote Sensing}, 2021.

\bibitem[{Seelbach Benkner} et~al.(2021){Seelbach Benkner}, {L\"{a}hner},
  {Golyanik}, {Wunderlich}, {Theobalt}, and {Moeller}]{SeelbachBenkner2021}
Marcel {Seelbach Benkner}, Zorah {L\"{a}hner}, Vladislav {Golyanik}, Christof
  {Wunderlich}, Christian {Theobalt}, and Michael {Moeller}.
\newblock Q-match: Iterative shape matching via quantum annealing.
\newblock In \emph{International Conference on Computer Vision (ICCV)}, 2021.

\bibitem[{Seelbach Benkner} et~al.(2023){Seelbach Benkner}, {Krahn},
  {Tretschk}, {L{\"a}hner}, {Moeller}, and {Golyanik}]{SeelbachBenkner2023}
Marcel {Seelbach Benkner}, Maximilian {Krahn}, Edith {Tretschk}, Zorah
  {L{\"a}hner}, Michael {Moeller}, and Vladislav {Golyanik}.
\newblock {QuAnt: Quantum Annealing with Learnt Couplings}.
\newblock \emph{International Conference on Learning Representations (ICLR)},
  2023.

\bibitem[Shi et~al.(2020)Shi, Tang, and min Jin]{Shi2020TrainingAQ}
Ruoxi Shi, Hao Tang, and Xian min Jin.
\newblock Training a quantum pointnet with nesterov accelerated gradient
  estimation by projection.
\newblock In \emph{Advances in Neural Information Processing Systems (NeurIPS)
  Workshops}, 2020.

\bibitem[Sim et~al.(2019)Sim, Johnson, and Aspuru-Guzik]{Sim_2019}
Sukin Sim, Peter~D. Johnson, and Al\'{a}n Aspuru-Guzik.
\newblock Expressibility and entangling capability of parameterized quantum
  circuits for hybrid quantum-classical algorithms.
\newblock \emph{Advanced Quantum Technologies}, 2019.

\bibitem[Srikumar et~al.(2021)Srikumar, Hill, and Hollenberg]{Srikumar:2021yzo}
Maiyuren Srikumar, Charles~D Hill, and Lloyd~CL Hollenberg.
\newblock Clustering and enhanced classification using a hybrid quantum
  autoencoder.
\newblock \emph{Quantum Science and Technology}, 2021.

\bibitem[Tan et~al.(2018{\natexlab{a}})Tan, Gao, Lai, and Xia]{8578710}
Qingyang Tan, Lin Gao, Yu-Kun Lai, and Shihong Xia.
\newblock Variational autoencoders for deforming 3d mesh models.
\newblock In \emph{Computer Vision and Pattern Recognition (CVPR)},
  2018{\natexlab{a}}.

\bibitem[Tan et~al.(2018{\natexlab{b}})Tan, Gao, Lai, Yang, and
  Xia]{tan2018mesh}
Qingyang Tan, Lin Gao, Yu-Kun Lai, Jie Yang, and Shihong Xia.
\newblock Mesh-based autoencoders for localized deformation component analysis.
\newblock In \emph{AAAI Conference on Artificial Intelligence},
  2018{\natexlab{b}}.

\bibitem[Tian et~al.(2022)]{https://doi.org/10.48550/arxiv.2206.03066}
Jinkai Tian et~al.
\newblock Recent advances for quantum neural networks in generative learning.
\newblock \emph{arXiv e-prints}, 2022.

\bibitem[Tretschk et~al.(2020)Tretschk, Tewari, Zollh\"{o}fer, Golyanik, and
  Theobalt]{Tretschk2020DEMEA}
Edgar Tretschk, Ayush Tewari, Michael Zollh\"{o}fer, Vladislav Golyanik, and
  Christian Theobalt.
\newblock {{DEMEA}: Deep Mesh Autoencoders for Non-Rigidly Deforming Objects}.
\newblock \emph{European Conference on Computer Vision (ECCV)}, 2020.

\bibitem[Wang et~al.(2023)Wang, Huang, Liu, Yi, Wu, and Wang]{e25030427}
Maida Wang, Anqi Huang, Yong Liu, Xuming Yi, Junjie Wu, and Siqi Wang.
\newblock A quantum-classical hybrid solution for deep anomaly detection.
\newblock \emph{Entropy}, 2023.

\bibitem[Yan et~al.(2022)Yan, Yang, Li, Guan, Kang, Hua, and
  Huang]{yan2022implicit}
Siming Yan, Zhenpei Yang, Haoxiang Li, Li~Guan, Hao Kang, Gang Hua, and Qixing
  Huang.
\newblock Implicit autoencoder for point cloud self-supervised representation
  learning.
\newblock \emph{arXiv preprint arXiv:2201.00785}, 2022.

\bibitem[Yang et~al.(2018)Yang, Feng, Shen, and Tian]{yang2018foldingnet}
Yaoqing Yang, Chen Feng, Yiru Shen, and Dong Tian.
\newblock Foldingnet: Point cloud auto-encoder via deep grid deformation.
\newblock In \emph{Computer Vision and Pattern Recognition (CVPR)}, 2018.

\bibitem[Yang and Sun(2022)]{Yang_2022_CVPR}
Yuan-Fu Yang and Min Sun.
\newblock Semiconductor defect detection by hybrid classical-quantum deep
  learning.
\newblock In \emph{Computer Vision and Pattern Recognition (CVPR)}, 2022.

\bibitem[{Yen-Chi Chen} et~al.(2021){Yen-Chi Chen}, {Wei}, {Zhang}, {Yu}, and
  {Yoo}]{2021arXiv210106189Y}
Samuel {Yen-Chi Chen}, Tzu-Chieh {Wei}, Chao {Zhang}, Haiwang {Yu}, and Shinjae
  {Yoo}.
\newblock {Hybrid Quantum-Classical Graph Convolutional Network}.
\newblock \emph{arXiv e-prints}, 2021.

\bibitem[Yurtsever et~al.(2022)Yurtsever, Birdal, and Golyanik]{Yurtsever2022}
Alp Yurtsever, Tolga Birdal, and Vladislav Golyanik.
\newblock Q-fw: A hybrid classical-quantum frank-wolfe for quadratic binary
  optimization.
\newblock In \emph{European Conference on Computer Vision (ECCV)}, 2022.

\bibitem[Zaech et~al.(2022)Zaech, Liniger, Danelljan, Dai, and
  Van~Gool]{Zaech_2022_CVPR}
Jan-Nico Zaech, Alexander Liniger, Martin Danelljan, Dengxin Dai, and Luc
  Van~Gool.
\newblock Adiabatic quantum computing for multi object tracking.
\newblock In \emph{Computer Vision and Pattern Recognition (CVPR)}, 2022.

\bibitem[Zhang et~al.(2021)Zhang, Kong, Farooq, Yung, Guo, and
  Wang]{Zhang_2021}
Xiao-Ming Zhang, Weicheng Kong, Muhammad~Usman Farooq, Man-Hong Yung, Guoping
  Guo, and Xin Wang.
\newblock Generic detection-based error mitigation using quantum autoencoders.
\newblock \emph{Physical Review A}, 2021.

\bibitem[Zhou et~al.(2020)Zhou, Wu, Li, Cao, Ye, Saragih, Li, and
  Sheikh]{Zhou2020}
Yi~Zhou, Chenglei Wu, Zimo Li, Chen Cao, Yuting Ye, Jason Saragih, Hao Li, and
  Yaser Sheikh.
\newblock Fully convolutional mesh autoencoder using efficient spatially
  varying kernels.
\newblock In \emph{Advances in Neural Information Processing Systems
  (NeurIPS)}, 2020.

\bibitem[Zhu et~al.(2021)Zhu, Bai, Wang, Li, and Chiribella]{Zhu:2021jnc}
Yan Zhu, Ge~Bai, Yuexuan Wang, Tongyang Li, and Giulio Chiribella.
\newblock {Quantum autoencoders for communication-efficient quantum cloud
  computing}.
\newblock \emph{arXiv e-prints}, 2021.

\end{thebibliography}

\newpage
\begin{center}
\textbf{\Large \color{bmvcblue}{Appendix (3D-QAE)}}
\end{center}
\appendix

This appendix provides extensive background on gate-based quantum computing (Sec.~\ref{sec:background}), mathematical details on our normalisation scheme (Sec.~\ref{sec:normalization}) and a visualisation of the circuit design (Sec.~\ref{sec:circuit_visualization}). 

\section{Background}\label{sec:background}

\subsection{Preliminaries on Quantum Computing}\label{sec:preliminaries}

A \emph{qubit} is the unit of information in a quantum system. 
Like a classical bit's binary states, a qubit has two basis states written in the Dirac's bra-ket notation as $\ket{0}$ and $\ket{1}$. 

\begin{wrapfigure}{r}{0.28\columnwidth}
    \vspace{4pt}
    \includegraphics[width=0.25\columnwidth]{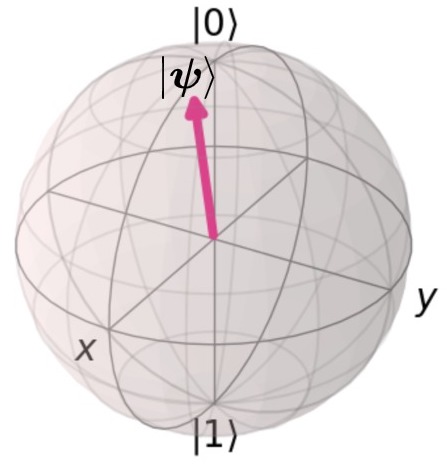}
     \caption{The one-qubit state $\ket{\psi}=(-0.91-0.39j)\ket{0}$ $+(-0.11{-}0.05j)\ket{1}$ on the Bloch sphere.} %
     \vspace{-17pt}
     \label{fig:bloch}
 \end{wrapfigure}

\emph{Superposition} is one of the key advantages of quantum computing over classical computing. 
Rather than being restricted to the basis states, a qubit can also be in a state $\ket{\psi} $ that is a linear combination of the basis states: $\ket{\psi} = \alpha \ket{0} + \beta 
\ket{1}$, where $\alpha,\beta\in\mathbb{C}$ and $\abs{\alpha}^2 + \abs{\beta}^2 = 1$. 
If we write the basis states as orthogonal vectors $\ket{0} = \begin{bmatrix}1&0\end{bmatrix}^T\in\mathbb{C}^2$ and $\ket{1} = \begin{bmatrix}0&1\end{bmatrix}^T\in\mathbb{C}^2$, we see that they span a complex Hilbert space. %
A state of a single qubit can be visualised on the \emph{Bloch sphere}; see Fig.~\ref{fig:bloch}. %

\emph{Measuring} the state $\ket{\psi}$ irreversibly forces (or \emph{collapses}) it into one of the basis states, \textit{i.e.} it yields either $\ket{0}$ or $\ket{1}$. 
The probability of collapsing to $\ket{0}$ or $\ket{1}$ is $|\alpha|^2$ and $|\beta|^2$, respectively. 
Because of this property, $|\alpha|$ and $|\beta|$ are also called \emph{probability amplitudes}.

\emph{Quantum entanglement} is the second key advantage over classical computing. 
The state of a collection (or \emph{system}) of classical bits is described fully by knowing the state of each bit. 
However, qubits can be so strongly correlated that the state of the system cannot be described anymore as a mere collection of per-qubit states. 
Rather, the entire system can be in a joint superposition.
For example, a system of two entangled qubits has a state $\ket{\psi} = \alpha \ket{00} + \beta \ket{01} + \gamma \ket{10} + \delta \ket{11}$, with $\alpha,\beta,\gamma,\delta\in\mathbb{C}$, $\abs{\alpha}^2 + \abs{\beta}^2 +\abs{\gamma}^2 +\abs{\delta}^2 = 1$, and $\ket{ij}\in\mathbb{C}^{2}\otimes\mathbb{C}^{2}$ are basis states covering all combinations of the two qubits, where ``$\otimes$'' is the tensor product. 
In other words, the state of a classical system is a \emph{single} combination of $N$ bits, while the state of an entangled system is a \emph{distribution} over \emph{all} combinations. 
More generally, an $N$-qubit system can exist in any superposition of the $2^N$ basis states: $\ket{\psi} = \sum_{i=0}^{2^N-1} \alpha_i \ket{i}$, where $i$ enumerates all combinations of the $N$ qubits (\textit{i.e.,} all basis states $\ket{i}\in(\mathbb{C}^{2})^{\otimes N}$%
) and  $\sum_{i=0}^{2^N-1} |\alpha_i|^2 = 1$. 
In vector notation, we obtain the \emph{state vector}: $\ket{\psi}= (\alpha_0,\alpha_1,\ldots,\alpha_{2^N-1})$. 
Thus, the state of $N$ qubits is given by specifying $2^N{-}1$ many degrees of freedom (DoF), while a classical system is given by $N$ DoF. 
Therefore, entanglement allows to encode exponentially many real numbers in $N$ many qubits, improving the processing speed of quantum computers and achieving exponential speed-up over classical systems.

\subsection{Quantum Circuits} 
\begin{figure} %
\centering
    \includegraphics[width=\columnwidth]{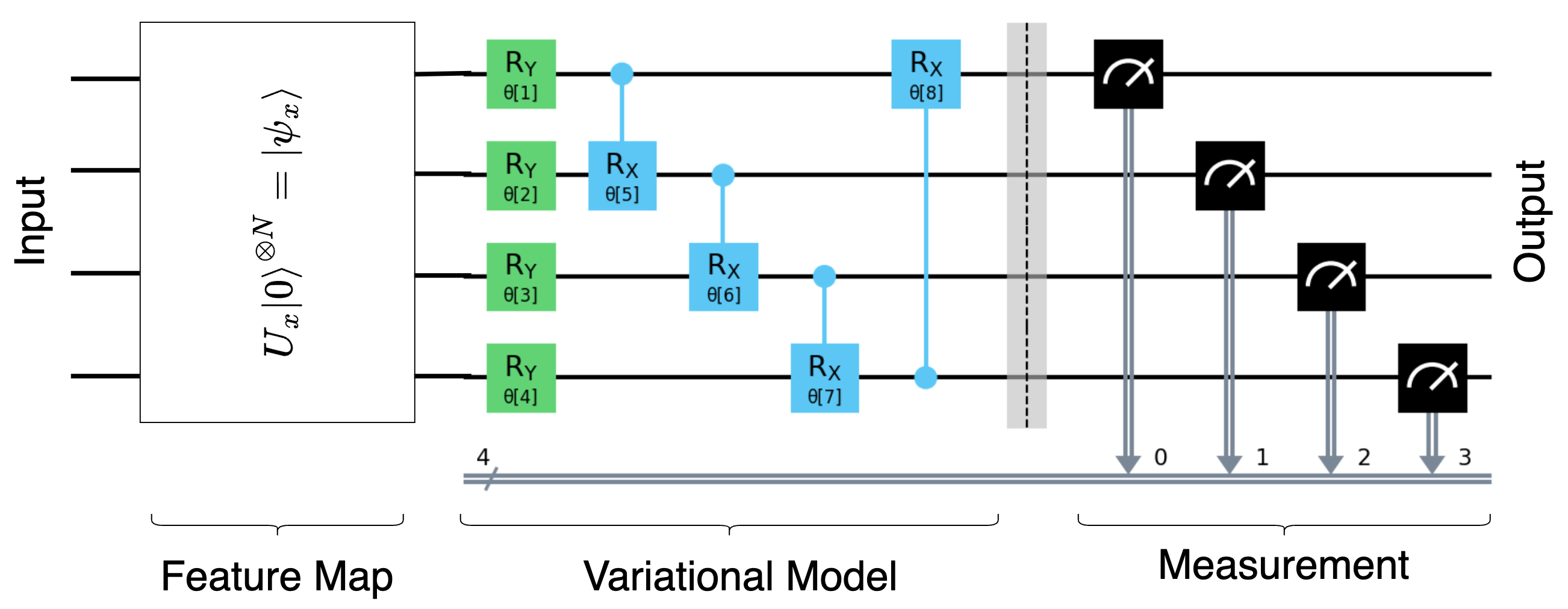}
    \caption{An exemplary quantum circuit. A feature map encodes a classical input into a state vector of $N{=}4$ qubits, which is then transformed by the parametrised quantum gates before being measured to obtain $N$ classical output bits. 
    } 
    \label{fig:quantu_circuit_background}
\end{figure}

Like a classical circuit acts on classical bits, a \emph{quantum circuit} transforms the state of a given $N$-qubit system (performs a computation with qubits). 
A quantum circuit consists of three broad steps, as Fig.~\ref{fig:quantu_circuit_background} shows: input encoding, applying parameterised quantum gates, and output measuring. 
First, the classical input data (3D poses in our case) needs to be encoded into an initial $N$-qubit state vector.
As its main operation, the quantum circuit then applies a unitary operation (the complex analogue of an orthogonal matrix) to this initial state vector. 
We thus obtain a transformed state vector as output. 
Subsequent measuring collapses the qubits to basis states, yielding an $N$-dimensional bit string. 

\noindent\textbf{Input Encoding.} 
We first need to convert our classical input data into an initial quantum state vector. 
To that end, a \emph{feature map} takes the classical input into the $N$-qubit Hilbert space.
While there are many ways of encoding classical information (\textit{e.g.}~angle encoding), we choose to use \emph{amplitude encoding} since it takes the most advantage of the exponentially large state space. %
It encodes $2^N$ distinct floating-point values as the amplitudes of the state, thereby requiring only $N$ qubits. %
Specifically, amplitude encoding takes a classical data vector 
$\mathbf{x} = (x_0,x_1,\ldots,x_{2^N-1})\in[0,1]^{2^N}$ that is normalised to $\lVert \mathbf{x}\rVert_2=1$ and encodes it into the $N$-qubit quantum state $\ket{\psi} = \sum_{i=0}^{2^N-1} (x_i + 0j)\ket{i}$, where $j$ is the imaginary unit.

\noindent\textbf{Parametrised Quantum Gates.} 
At its core, a quantum circuit applies a unitary transform $U_\theta\in\mathbb{C}^{2^N\times 2^N}$ (with parameters $\theta$) to this initial state vector. 
In practice, $U$ is implemented by sequentially applying \emph{quantum gates}: $U=U_G \cdots U_2 U_1$. 
A quantum gate implements a simple unitary transform $U_k$ that maps a state vector $\ket{\psi}$ to a new state vector $\ket{\phi} = U_k\ket{\psi}$. 
During training, we optimise for the best set of parameters $\theta$ of these gates. 

We next discuss \emph{hardware-efficient} gates that can be directly realised in quantum hardware. 
The parameter-free $I$, $X$-, $Y$-, and $Z$-Pauli gates act on a single qubit. 
The $I$ gate is an identity map, while the $X$, $Y$, and $Z$ gates rotate the qubit by $180\degree$ around the corresponding axis. 
The parametrised Pauli rotation gates $R_X, R_Y, R_Z$ rotate the qubit by a given angle around the corresponding axis. %
For example, $R_X(\theta{=}20\degree)$ rotates the qubit by $\theta{=}20\degree$ around the $X$-axis, where $\theta$ is the parameter of the gate. %

The controlled rotation gates $\mathit{CR}_X$, $\mathit{CR}_Y$, and $\mathit{CR}_Z$ work on two qubits and they thus modify the entanglement of these qubits. 
One qubit acts as the \emph{control qubit}: If it is in state $\ket{0}$, then the identity map is applied to the other qubit (called the \emph{target qubit}); and if it is in state $\ket{1}$, then $R_X$, $R_Y$, or $R_Z$ are applied to the target qubit. 
Importantly, if the control qubit is in a superposition, both operations are applied to the target qubit according to the superposition. 

\begin{figure}
\centering
    \includegraphics[width=\columnwidth]{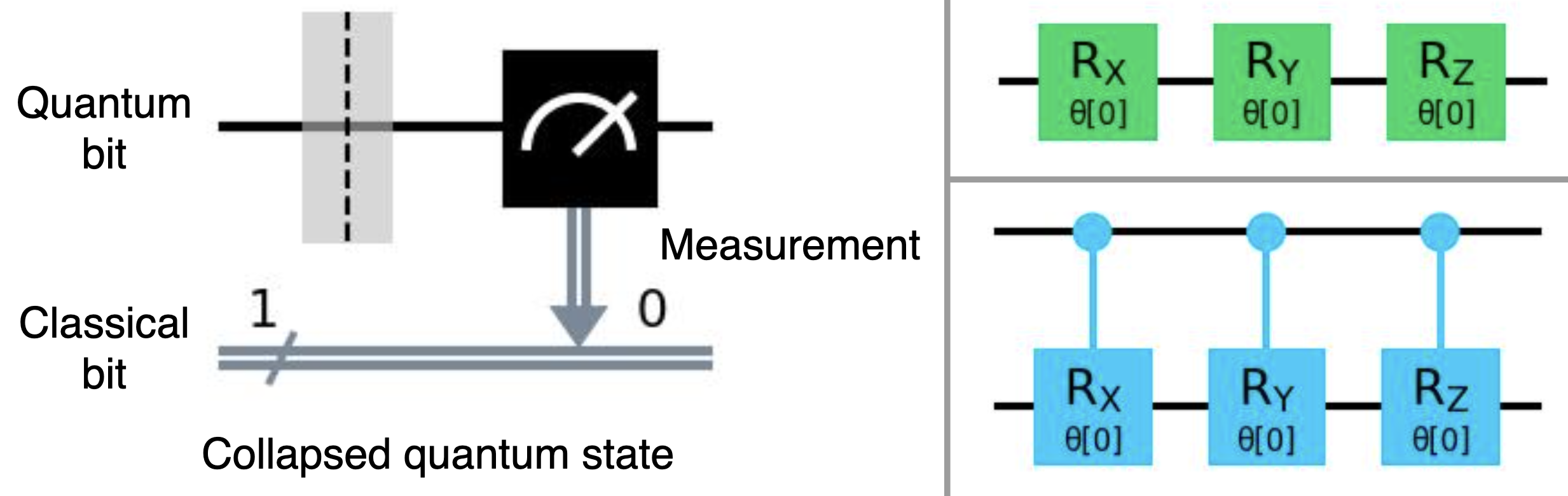}
    \caption{Circuit Notation. 
    (Left:) Measurement: The black wire is a qubit and the grey double wire is a classical bit; the black box indicates measurement. 
    (Top-right:) Rotation gates. 
    (Bottom-right:) 
    Controlled rotation gates. %
    The top qubit is the control qubit for all three gates, while the bottom qubit acts as the target qubit. 
    }
    \label{fig:notation}
\end{figure}

Fig.~\ref{fig:notation} shows the notation of these gates. 
Note that they form a \emph{universal} set of gates \cite{ NielsenChuang2000}: They are enough to approximate \emph{any} unitary transformation arbitrarily well as a finite sequence of them.
A gate can occur more than once in the sequence and can be applied to any of the qubits each time. 
We also note that since gates are unitary transforms, they are reversible. 
In fact, each of these gates happens to be its own inverse (\textit{e.g.}~$(\mathit{CR}_X(20\degree))^{\dagger} = \mathit{CR}_X(-20\degree)$). %

\noindent\textbf{Output Measurement.} 
Lastly, we measure each of the $N$ qubits in a particular computational basis (typically along the $Z$-axis, which is also the basis we use in Sec.~\ref{sec:preliminaries}), collapsing the state vector into a binary vector of length $N$.

\subsection{Circuit Design and Barren Plateaus} 
One widespread challenge when optimising the parameters $\theta$ of a quantum circuit are barren plateaus \cite{McClean_2018}. %
In some cases, the quantum nature of the task considered induces a particular circuit design and parameter choices, \textit{e.g.}~in quantum chemistry.
Unfortunately, this approach is not feasible for generic tasks that are not inherently quantum-related. 
Instead, we need to follow the heuristic approach of designing generic circuits that can be efficiently implemented in hardware. 
In this heuristic setting, we usually follow a classical gradient-based optimisation routine that uses a cost function to iteratively compute updates to the parameters. 
Crucially, using more and more gates or qubits in generic designs leads to a loss landscape that is virtually flat (a barren plateau) in most places. 
The resulting per-parameter derivatives are essentially random, with smaller and smaller magnitudes and variances. 
These uninformative, vanishing gradients hinder the optimisation and thus restrict the size of the quantum circuits, limiting the expressibility. %

\subsection{Optimising Quantum Circuits} 
A further challenge unique to quantum circuits is that gradient-based optimisation inherently scales badly on real quantum hardware. 
Ultimately, this is because measurement irreversibly collapses the state. 
Thus, %
the \emph{parameter shift} update rule requires evaluating the loss twice per parameter, which scales much worse than, for example, back-propagation on classical hardware. 
We avoid this issue in practice by resorting to simulation on classical hardware. 
This allows us to recover from the measurement process without having to re-run the circuit and we can thus apply back-propagation. 
In addition, simulation avoids noise, which remains prominent in contemporary quantum hardware.

\section{Normalisation Scheme}
\label{sec:normalization}

The dataset is a set of $N$ classical 3D point clouds $[\{\mathbf{v}^j_i\in\mathbb{R}^3\}^{V-1}_{i=0}]^{N-1}_{j=0}$, each with $V$ vertices. 
The probability vector containing the amplitudes restricts the output vector to the positive octant. 
To combat this, we take several steps to keep the raw input data within the positive octant as well. 

We first define an axis-aligned bounding box $(\mathbf{v}_{min},\mathbf{v}_{max})$ across the entire dataset. 
This is done by defining the minimum and maximum values along each axis: 
\begin{align}
    v_{min,a} &= \min_{\substack{j=0,\ldots,N-1\\ i=0,\ldots,V-1}} v^j_{i,a},\\ 
    v_{max,a} &= \max_{\substack{j=0,\ldots,N-1\\ i=0,\ldots,V-1}} v^j_{i,a}, 
\end{align} 
where $v^j_{i,a}\in\mathbb{R}$ is the coordinate of vertex $\mathbf{v}^j_{i}$ along axis $a\in\{x,y,z\}$. 

To achieve isotropic re-scaling, we turn the bounding box into a cube with side length $\mathbb{R}\ni s=\max_{a\in{x,y,z}} v_{max,a} - v_{min,a}$. 
We first shift the data and then re-scale it to get the final normalised dataset:
\begin{equation}
\bigg[\{\mathbf{\tilde{v}}^j_i\}^{V-1}_{i=0}\bigg]^{N-1}_{j=0} = \bigg[\biggl\{\frac{\mathbf{v}^j_i-\mathbf{v}_{min}}{s}\biggr\}^{V-1}_{i=0}\bigg]^{N-1}_{j=0}.
\end{equation}

\section{Circuit Design}
\label{sec:circuit_visualization}

\begin{figure}
    \centering
    \includegraphics[width=0.99\columnwidth]{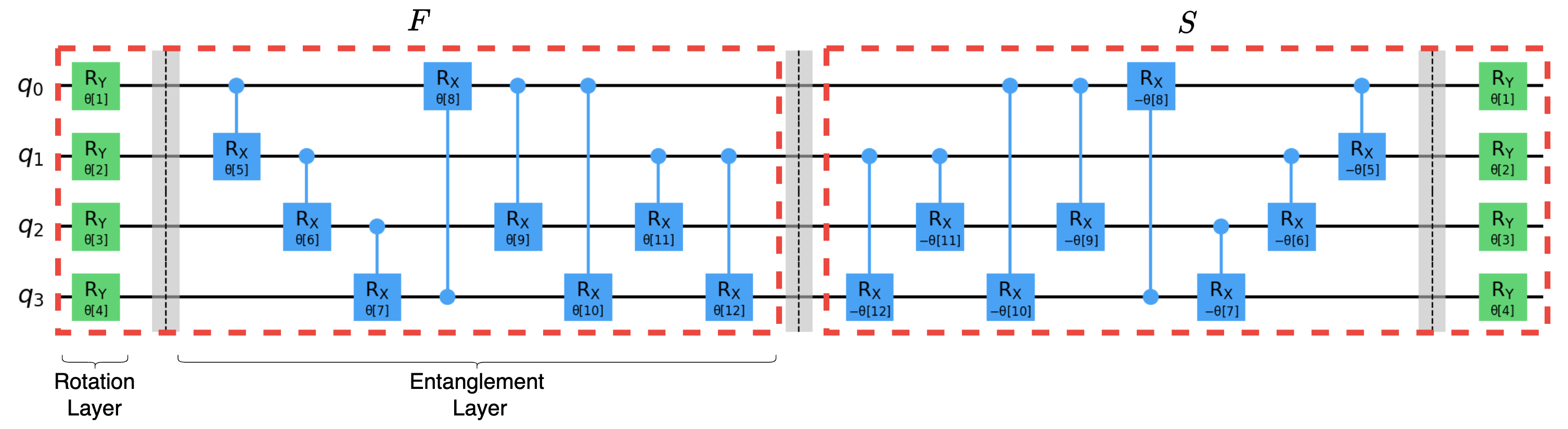}
\caption{``Inverse'' Scheme for Blocks.  
$F$, on the left, uses the basic block B architecture. 
Its inverse, $S$, is on the right. 
Here, $S$ uses the random initialisation. 
}
\label{fig:inverse_block}
\end{figure}

Fig.~\ref{fig:inverse_block} visualises the ``inverse'' architecture type. 

\end{document}